\documentclass[letterpaper, 10 pt, conference]{ieeeconf}  

\IEEEoverridecommandlockouts                              

\usepackage{graphics} 
\usepackage{amsmath} 
\usepackage{amssymb}  
\usepackage{booktabs}
\usepackage{siunitx}
\usepackage{multirow}
\usepackage{makecell}
\usepackage{subcaption} 
\usepackage[usenames, dvipsnames,table,xcdraw]{xcolor}
\usepackage{pifont}
\newcommand{\greencheckmark}{{\color{Green}\checkmark}}
\newcommand{\redxmark}{{\color{red}\ding{53}}}

\usepackage[capitalise]{cleveref}
\usepackage{bm}
\usepackage{mathtools}
\newcommand{\myM}[1]{\bm{\mathit{#1}}} 

\newcommand{\bfun}{\myM{f}}

\newcommand{\bq}{\myM{q}}
\newcommand{\btau}{\myM{\tau}}

\newcommand{\bu}{\myM{u}}
\newcommand{\bv}{\myM{v}}

\newcommand{\bx}{\myM{x}}

\newcommand{\mechname}  [1] {\textsc{#1}}
\newcommand{\rfig}                                      [1]     {Fig.~\ref{#1}}

\newcommand{\rtab}                                      [1]     {Table~\ref{#1}}
\newcommand{\rsec}                                      [1]     {Section~\ref{#1}}
\renewcommand{\cdot}{}

\title{\LARGE \bf
Design, analysis and control of the series-parallel hybrid RH5 humanoid robot
}

\author{Julian Esser$^{1}$, Shivesh Kumar$^{1}$, Heiner Peters$^{1}$, Vinzenz Bargsten$^{1}$, Jose de Gea Fernandez$^{1}$, Carlos Mastalli$^{2}$, \\ Olivier Stasse$^{3}$ and Frank Kirchner$^{1}$
\thanks{$^{1}$The authors are with the Robotics Innovation Center, DFKI GmbH, 28359 Bremen, Germany. Corresponding Author's Email:
        {\tt\small shivesh.kumar@dfki.de}}%
\thanks{$^{2}$Carlos Mastalli is with the Alan Turing Institute at the University of Edinburgh, Edinburgh, United Kingdom.}%
\thanks{$^{3}$Olivier Stasse is with GEPETTO group at LAAS-CNRS, Toulouse, France.}
\thanks{This research was supported by the German Aerospace Center (DLR) with federal funds (Grant Numbers: FKZ 50RA1701 and FKZ 01IW20004 respectively) from the Federal Ministry of Education and Research (BMBF). O. Stasse and C. Mastalli acknowledge the support of the European Commission under the  Horizon 2020 project Memory of Motion (MEMMO, project ID:  780684), and the Engineering  and  Physical Sciences Research Council (EPSRC) UK RAI Hub for  Offshore Robotics for Certification of Assets (ORCA, grant reference EP/R026173/1).}
}

\begin{document}

\maketitle
\thispagestyle{empty}
\pagestyle{empty}

\begin{abstract}
Last decades of humanoid research has shown that humanoids developed for high dynamic performance require a stiff structure and optimal distribution of mass--inertial properties.
Humanoid robots built with a purely tree type architecture tend to be bulky and usually suffer from velocity and force/torque limitations. This paper presents a novel series-parallel hybrid humanoid called RH5 which is 2 m tall and weighs only 62.5 kg capable of performing heavy-duty dynamic tasks with 5 kg payloads in each hand. The analysis and control of this humanoid is performed with whole-body trajectory optimization technique based on differential dynamic programming (DDP). Additionally, we present an improved contact stability soft-constrained DDP algorithm which is able to generate physically consistent walking trajectories for the humanoid that can be tracked via a simple PD position control in a physics simulator. Finally, we showcase preliminary experimental results on the RH5 humanoid robot.
\end{abstract}

\section{Introduction}
Humanoid robots are designed to resemble the human body and/or human behavior.
Recent research indicates that humanoid robots require a stiff structure and good mass distribution for high dynamic tasks \cite{Stasse2019}.
These properties can be easily achieved by utilizing Parallel Kinematic Mechanisms (PKM) in the design, as they provide higher stiffness, accuracy, and payload capacity compared to serial robots. However, most existing bipedal robot designs are based on serial kinematic chains.

Series--parallel hybrid designs combining the advantages of serial and parallel topologies are commonly used in the field of heavy machinery, e.g., cranes, excavator arms, etc. However, such designs also have recently caught the attention of robotics researchers from industry and academia~(see \cite{kumar2020survey} for an extensive survey).  
For instance, the \mechname{Lola} humanoid robot~\cite{2006_humanoids_lolalegdesign_lohmeier} has a spatial slider crank mechanism in the knee joint and a two DOF rotational parallel mechanism in the ankle joint. Similarly, the \mechname{Aila} humanoid robot~\cite{2011_AILA} employs parallel mechanisms for its wrist, neck, and torso joints. Furthermore, the design of the NASA \mechname{Valkyrie} humanoid robot~\cite{2015_nasa_valkyrie}, built by the NASA Johnson Space Center, follows a similar design concept by utilizing PKM modules for its wrist, torso and ankle joints. Both torque controlled humanoid robots \mechname{TORO} from DLR~\cite{Englsberger2014} and \mechname{TALOS}~\cite{2017_talos} from PAL Robotics mostly contain serial kinematic chains but utilize simple parallelogram linkages in their ankles for creating the pitch movement. 
The motivation of such hybrid designs is to achieve a lightweight and compact robot while enhancing the stiffness and dynamic characteristics.
However, the evaluation of the humanoid design is still non--trivial since it necessitates whole-body trajectory optimization techniques which exploit the full dynamics of the system. 

\begin{figure}
\begin{subfigure}{.16\textwidth}
	\includegraphics[width=.95\linewidth]{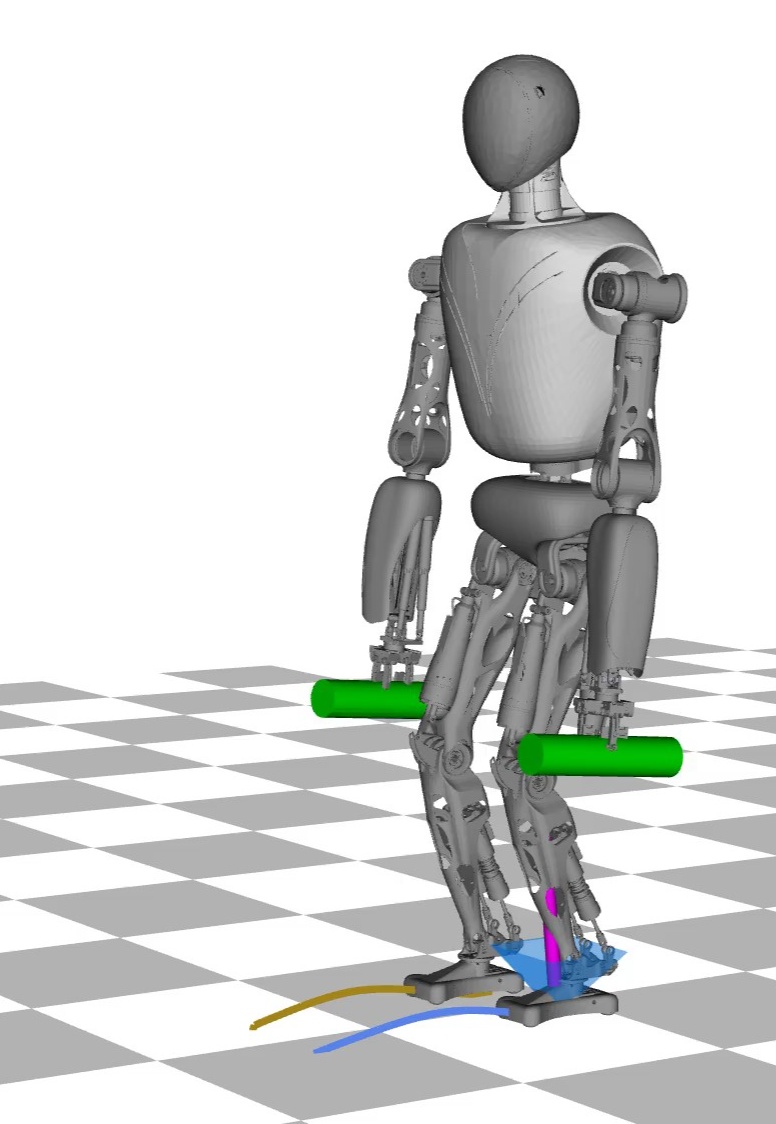}
\end{subfigure}%
\begin{subfigure}{.16\textwidth}
	\includegraphics[width=.95\linewidth]{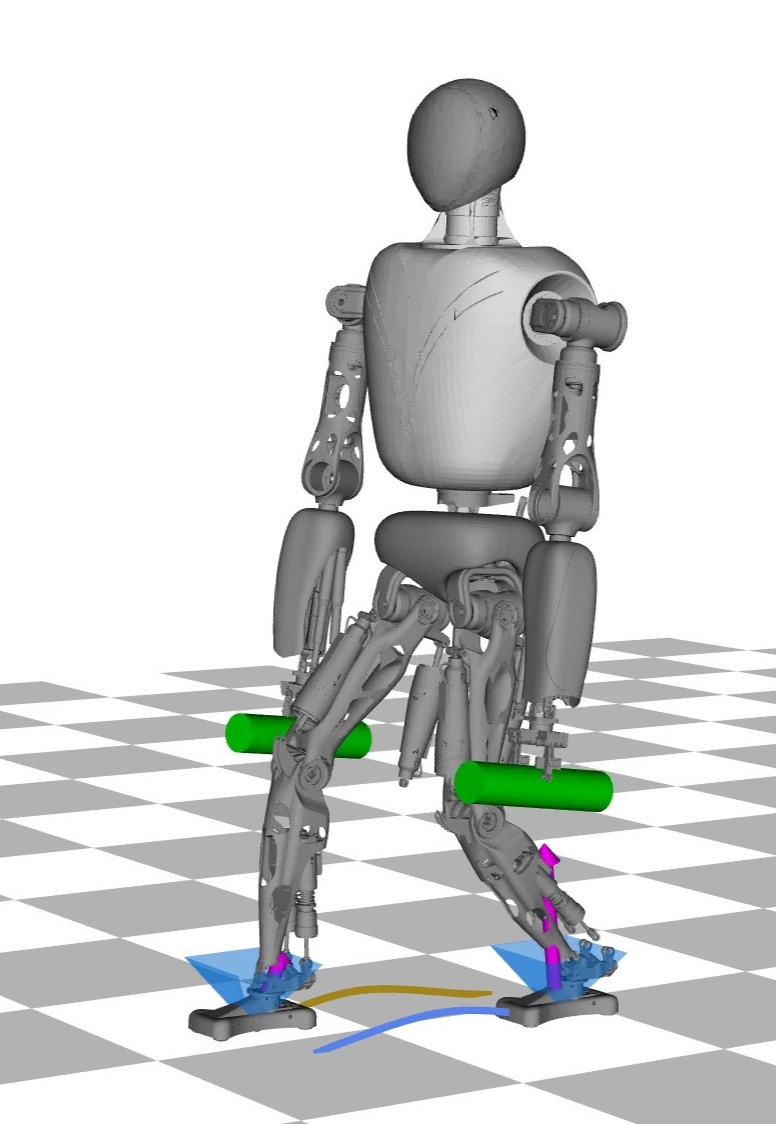}
\end{subfigure}%
\begin{subfigure}{.16\textwidth}
	\includegraphics[width=.95\linewidth]{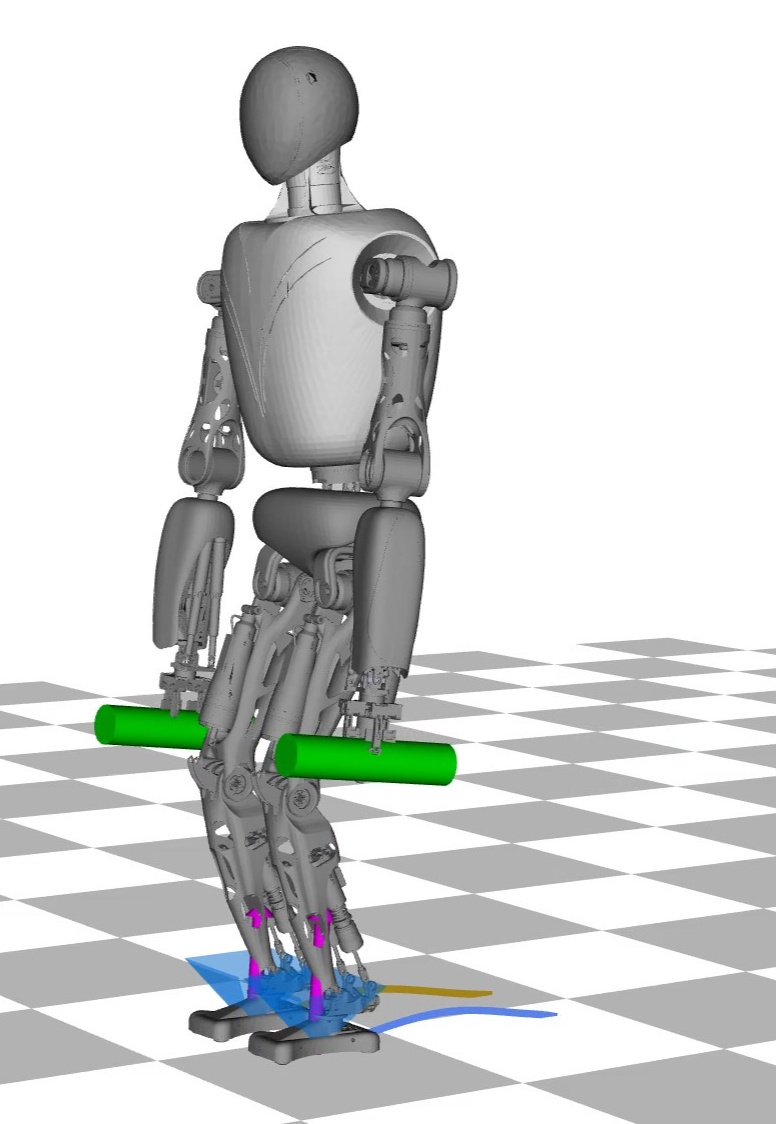}
\end{subfigure}%
\caption{The RH5 humanoid robot performing a dynamic walking motion while carrying two 5 kg bars.}
\label{fig:walkWeights_snaps}
\end{figure}

Trajectory Optimization (TO) is a numerical optimization technique that aims to find a state-control sequence, which locally minimizes a cost function and satisfies a set of constraints.
TO based on reduced centroidal dynamics~\cite{carpentier2016versatile, aceituno2017simultaneous} has become a popular approach in the legged robotics community. However, tracking of centroidal motions requires an instantaneous feedback linearization, where typically quadratic programs with task-space dynamics are solved (e.g., \cite{Englsberger2014}).
While TO based on reduced dynamics models has shown great experimental results (e.g.,~\cite{fahmi2019passive}), whole-body TO instead is proven to produce more efficient motions, with lower forces and impacts~\cite{budhiraja2018differential}.
To this end, we focus on a DDP~\cite{mayne1966} variant, called Box-FDDP \cite{mastalli2020direct}, to efficiently compute dynamic whole-body motions, as depicted in \cref{fig:walkWeights_snaps}. However, the trajectories generated with those solvers often require an additional stabilizing controller to reproduce the behavior in another simulator or the real robot~\cite{giraud2020motion}.

\begin{figure*}[!htb]
\centering	
\includegraphics[width=.95\textwidth]{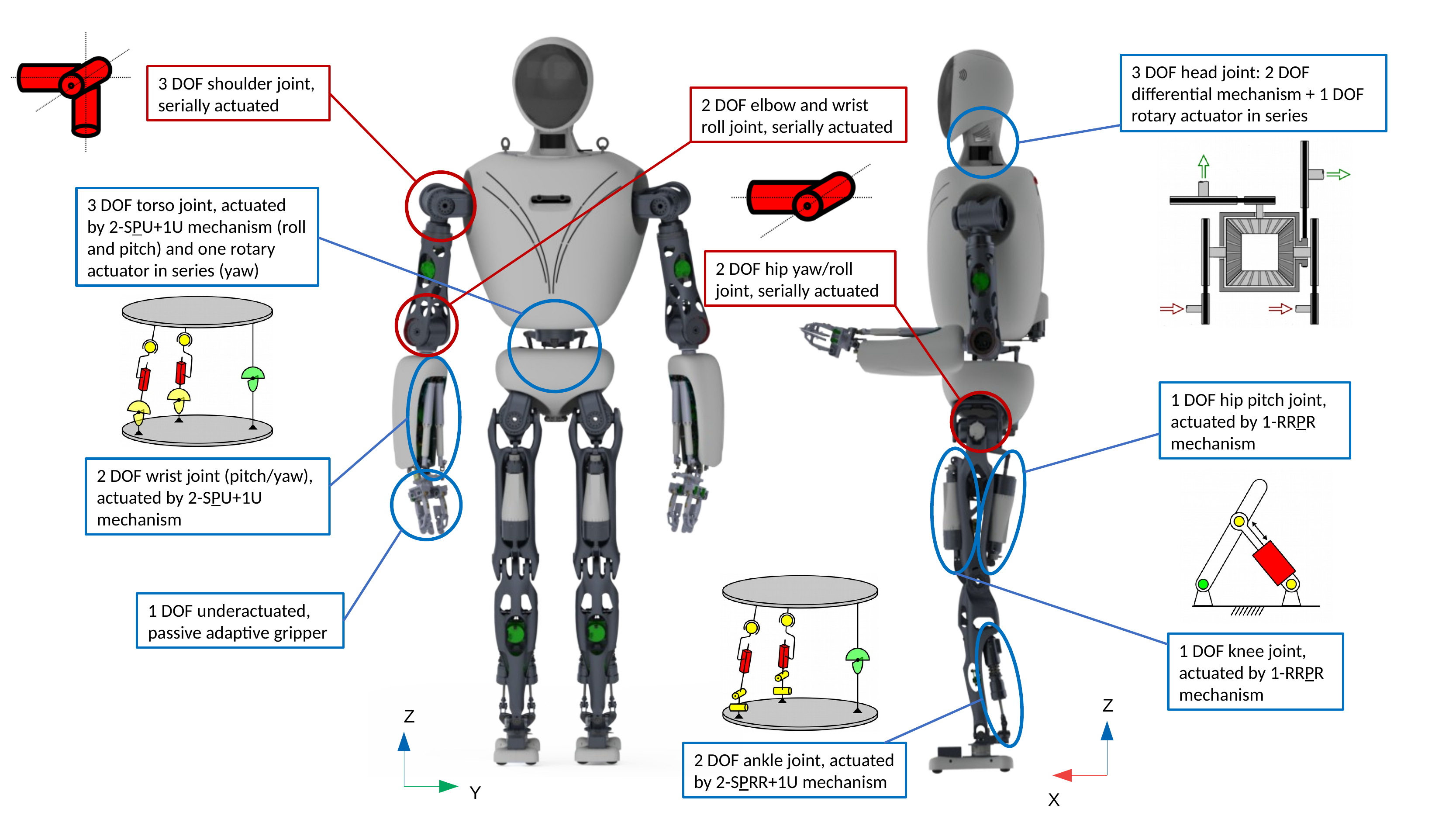}
\caption{Actuation and morphology of the RH5 humanoid robot (S: Spherical, R: Revolute, P: Prismatic, U: Universal)}
\label{fig:rh5_schematic}
\end{figure*}

\paragraph*{Contributions} First, we introduce RH5: a novel series--parallel hybrid humanoid robot that has a lightweight modular design, high stiffness and outstanding dynamic properties. Our robot can perform heavy-duty tasks and dynamic motions. Second, we present an analysis of the RH5 design by generating highly dynamic motions using the Box-FDDP algorithm. Third, we present a contact stability soft-constrained DDP trajectory optimization approach which generates physically consistent walking trajectories. Fourth, we present both simulation and preliminary experimental results on the RH5 robot.

\paragraph*{Organization}
 \rsec{sec_mechatronic_system_design_rh5} describes the mechatronic system design of the novel RH5 humanoid robot with details about its mechanical design, electronics design and processing architecture. \rsec{sec_analysis_and_control_using_DDP} presents the analysis and control of the system based on the Box-FDDP algorithm. \rsec{sec_results_discussions} presents the simulation and first experimental results on the system and \rsec{sec_conclusion} concludes the paper.

\section{System design of RH5 humanoid}
This section provides details on the mechanical design, electronics design and processing architecture of the RH5 humanoid robot. \label{sec_mechatronic_system_design_rh5}

\subsection{Mechanical Design}
The robot has been designed with proportions close to human. The robot has 34 DOF as depicted in \rfig{fig:rh5_schematic}. The robot is symmetric around the XZ plane, and its overall weight and height are 62.5 kg and 2 m, respectively. The RH5 robot has a series-parallel hybrid actuation that reduces its weight and improves its structural stiffness and dynamic characteristics. Below, we describe the actuation principle and design of legs, torso, head and arms.

\subsubsection{Actuation Principle}
We use serially arranged rotary actuators to increase the range of motion. However, for joints with small range of motion, we exploit the advantages of parallel kinematics. These include non--linear transmission ratio, superposition of forces of parallel actuators, higher joint stiffness and optimal mass distribution in order to reduce the inertia of the robot's extremities. 

We use high torque BLDC motors and harmonic drive gears for joints with direct rotary actuation in serial chains. We utilize this type of drive unit in the three DOF shoulder joints, torso (yaw), hip joints (yaw, roll), elbow and wrist (roll). 
The head joints are actuated with commercially available servo drives. Parallel drive concepts are implemented using linear drive units consisting of a high torque BLDC motor in combination with a ball screw. 
We actuate the hip joints (pitch), the body joint (pitch, roll) as well as the knee and ankle joints of the RH5 robot according to this design (see \rtab{tbl:rh5_range_of_motion_actuation} for an overview). Commercial linear drive units are used to actuate the wrists. Non-linear transmission of the parallel mechanisms was optimized and exploited especially in the joints for the forward movement of the locomotive extremities (hip pitch, knee, ankle pitch). The joint angle under which the highest torque occurs was chosen in such a way that it is within the range of the highest torque requirements to be expected according to gait pattern described in~\cite{Zoss2006}. Near the limits of the joint's Range Of Motion (ROM), the available torque decreases in favor of a higher speed. Using a highly integrated 2-S\underline{P}RR+1U parallel mechanism~\cite{2019_Kumar_RH5Ankle} in the lower extremities enables an ankle design that outperforms the ankle of similar humanoid robots at almost half of their weight (see \rtab{tbl:rh5_comparison_other_robots}).
\rtab{tbl:rh5_range_of_motion} shows the ROM, speed and torque limits in the generalized coordinates (see~\cite{kumar2019modular} for a detailed analysis).
\begin{table}[!t]
\centering
\begin{tabular}{ccccc}
	\toprule[0.03cm]
  Actuator & ROM ($\si{\milli\meter}$) & Max. force ($\si{\newton}$)  & Max. vel. ($\si{\milli\meter}/\si{\second}$) \\ 
   \midrule
   Wrist & $235$--$290$ &  $\num{495}$  & $\num{38}$ \\  
   Torso & $195$--$284$ &  $\num{2716}$  & $\num{291}$ \\  
   Hip3 & $272$--$431$ & $\num{4740}$ & $\num{175}$ \\ 
   Knee & $273$--$391$ &  $\num{5845}$  & $\num{140}$ \\ 
   Ankle & $221$--$331$ & $\num{2000}$ & $\num{265}$ \\ 
   \bottomrule
\end{tabular}
\caption{ROM of linear actuators of the RH5 robot.}
\label{tbl:rh5_range_of_motion_actuation}
\end{table}
\begin{table}[!t]
\centering
\begin{tabular}{ccccccc}
	\toprule[0.03cm]
  \makecell{Robot} & \makecell{Mass \\($\si{\kilo\gram}$)} & \makecell{Ankle \\DOF} & \makecell{ROM \\($^\circ$)} & \makecell{Torque \\($\si{Nm}$)}  & \makecell{Velocity \\$(^\circ/s)$} \\ 
   \midrule
   \multirow{2}{*}{TORO} & \multirow{2}{*}{7.65} & Roll & $-19.5$--$19.5$ &  $\num{40}$  & $\num{120}$ \\  
    & &Pitch & $-45$--$45$ &  $\num{130}$  & $\num{176}$ \\  
   \midrule
   \multirow{2}{*}{TALOS} & \multirow{2}{*}{6.65} & Roll & $-30$--$30$ &  $\num{100}$  & $\num{275}$ \\  
    & &Pitch & $-75$--$45$ &  $\num{160}$  & $\num{332}$ \\
    \midrule
   \multirow{2}{*}{RH5} & \multirow{2}{*}{3.6} & Roll & $-57$--$57$ &  $\num{84}$--$\num{158}$  & $\num{386}$--$\num{726}$ \\  
    & &Pitch & $-51.5$--$45$ &  $\num{121}$--$\num{304}$  & $\num{200}$--$\num{502}$ \\    
   \bottomrule
\end{tabular}
\caption{Comparison of lower limbs design characteristics between the TORO, TALOS and RH5 humanoid robots.}
\label{tbl:rh5_comparison_other_robots}
\end{table}

\subsubsection{Leg}
The two legs of the robot are identical in construction and follow a Spherical--Revolute--Universal (SRU) kinematic design. Each leg has a 3 DOF hip joint (realized with 2 DOF serial mechanism and 1-RR\underline{P}R mechanism), 1 DOF knee joint (1-RR\underline{P}R mechanism) and a 2 DOF ankle joint (2-S\underline{P}RR+1U mechanism). The rotation axes of the hip joint intersect at a single point that is located at approximately half of the total height of the robot at 930 mm. The distance between both hip joints is 220 mm. To adjust the available range of motion, the first joint axis was tilted by 15 degrees with respect to the XY-plane of the robot. The lengths of the upper and lower leg are almost identical with lengths of 410 and 420 mm, respectively.. Upper and lower leg are connected by the knee joint. The ankle joint has two rotation axes that intersect the same point. The axis intersection point is 100 mm above the ground contact surface. Contact with the ground is made via 4 contact points, which span a support polygon with an area of 80 mm x 200 mm. The total mass of a leg is 9.8 kg, of which 6.2 kg are assigned to the thigh and hip joint, 2.3 kg to the lower leg, and 1.3 kg to the foot, respectively. 

\subsubsection{Torso and Head}
We use a spherical body joint with 3 DOF (a 2-SPU+1U unit) to expand the body ROM, which translates to i) the realization of more complex walking patterns, ii) the improvement of the robot balance, and iii) a larger manipulation space. The intersection point of the joint axes is at a height of 1140 mm above the foot contact area and it weights 4.8 kg. The body joint carries the torso, which contains most of the electronics and the battery of the robot and acts as a connecting structure between the robot's extremities. The torso weighs 21 kg in total.  The robot also has a head that serves as a sensor carrier for imaging and acoustic perception. This includes a joint with 3 DOF. The intersection point of the joint axes is at a height of 1800 mm above the foot contact area. The head weighs 3.3 kg and includes laser scanner, stereocamera, microphones, infrared camera and some processing units. 

\begin{table}[t!]
\centering
\begin{tabular}{ccccc}
	\toprule[0.03cm]
  Joint & ROM ($^\circ$) & Max. torque ($\si{\newton\meter}$)  & Max. vel. ($\si{\degree}/\si{\second}$) \\ 
   \midrule
   Shoulder1 & $-180^\circ$--$180^\circ$ & $\num{135}$ & $\num{210}$\\ 
   Shoulder2 & $-110^\circ$--$110^\circ$ & $\num{167}$ & $\num{131}$ \\ 
   Shoulder3 & $-180^\circ$--$180^\circ$ & $\num{135}$ & $\num{210}$ \\ 
   Elbow & $-125^\circ$--$125^\circ$ &  $\num{23}$  & $\num{413}$ \\ 
   Wrist Roll & $-180^\circ$--$180^\circ$ & $\num{18}$ & $\num{660}$ \\ 
   Wrist Pitch & $-46.8^\circ$--$46.8^\circ$ &  $\num{24}$--$\num{35}$  & $\num{60}$--$\num{106}$ \\  
   Wrist Yaw & $-39.6^\circ$--$57.6^\circ$ &  $\num{22}$--$\num{35}$  & $\num{62}$--$\num{100}$ \\       
   Torso yaw & $-40^\circ$--$40^\circ$ & $\num{23}$ & $\num{413}$ \\    
   Torso pitch & $-25^\circ$--$29^\circ$ & $\num{380}$--$\num{493}$ & $\num{184}$--$\num{238}$ \\ 
   Torso roll & $-36^\circ$--$36^\circ$ &  $\num{285}$--$\num{386}$  & $\num{208}$--$\num{400}$ \\  
   Hip1 & $-180^\circ$--$180^\circ$ & $\num{135}$ & $\num{210}$\\ 
   Hip2 & $-46^\circ$--$67^\circ$ & $\num{135}$ & $\num{210}$ \\ 
   Hip3 & $-17^\circ$--$72^\circ$ & $\num{357}$--$\num{540}$ & $\num{88}$--$\num{133}$ \\ 
   Knee & $0^\circ$--$88^\circ$ &  $\num{337}$--$\num{497}$  & $\num{94}$--$\num{139}$ \\ 
   Ankle pitch & $-51.5^\circ$--$45^\circ$ & $\num{121}$--$\num{304}$ & $\num{200}$--$\num{502}$ \\ 
   Ankle roll & $-57^\circ$--$57^\circ$ &  $\num{84}$--$\num{158}$  & $\num{386}$--$\num{726}$ \\ 
   \bottomrule
\end{tabular}
\caption{ROM of the RH5 humanoid robot in its independent joint space (generalized coord. when robot is fixed).}
\label{tbl:rh5_range_of_motion}
\end{table}

\subsubsection{Arm}
The robot is equipped with two manipulators. Each manipulator includes a 3 DOF shoulder joint, an 1 DOF elbow, a 3 DOF wrist (realized with a rotary actuator in series with 2-S\underline{P}U+1U mechanism) and a 1 DOF underactuated gripper. The intersection points of the shoulder joint axes have a distance of 640 mm between the right and left shoulder. The first axis is tilted forward by 14 degrees with respect to the XZ-plane of the robot to increase the manipulation area in front of the torso. The lengths of the upper and lower arms are 355 mm and 386 mm, respectively. Upper and lower arm are coupled by the elbow joint. The three joint axes of the wrist also form a common point of intersection. The end effector is a self-adaptive three-finger gripper, whose individual fingers are simultaneously actuated. The upper and lower arm including gripper weight 3.6 and 3.3 kg, respectively.\\

\subsection{Electronic Design and Processing Architecture}

\begin{figure}
\centering	
\includegraphics[width=1.0\columnwidth]{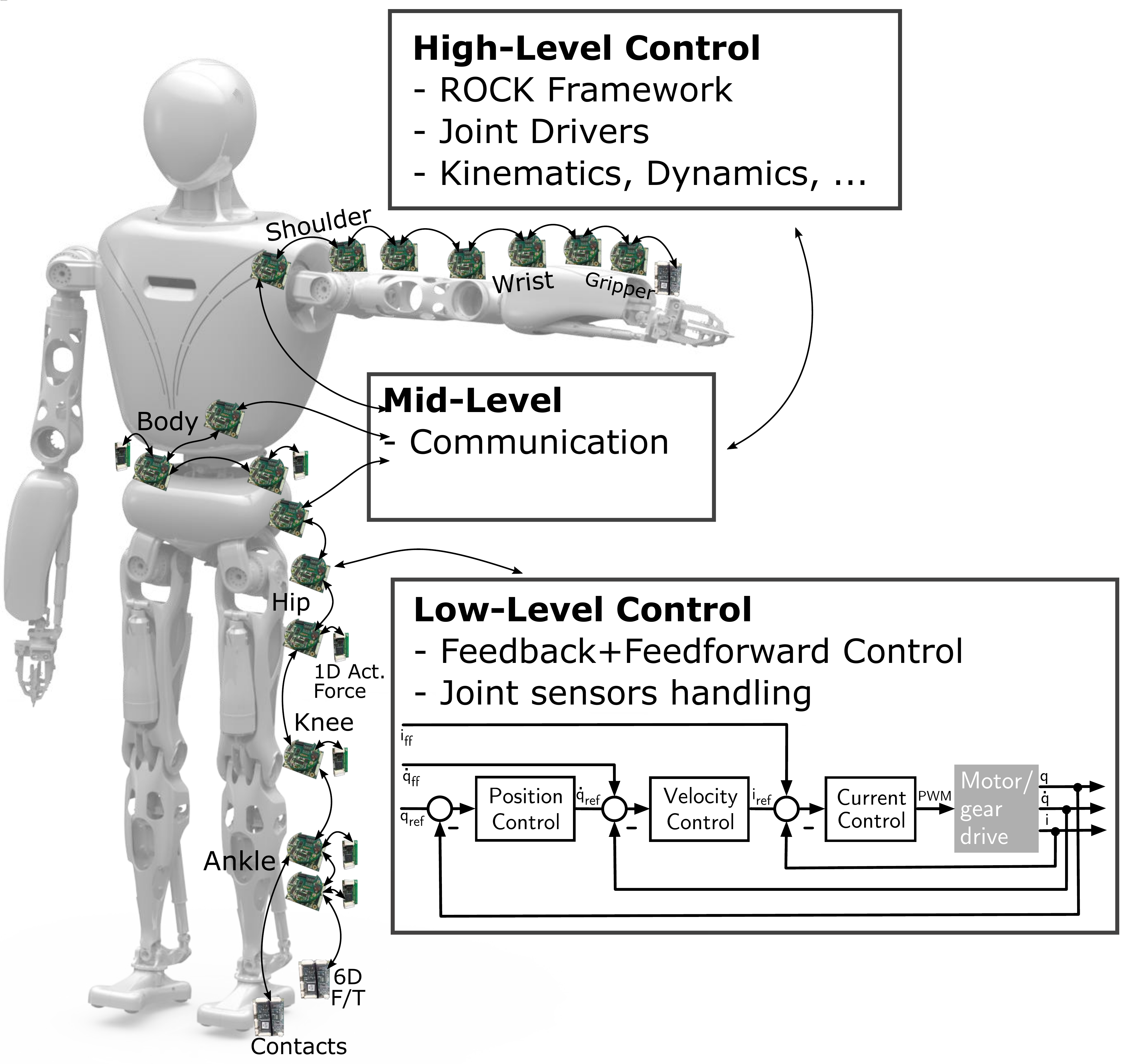}
\caption{Electronic and control units of the RH5 robot.}
\label{fig:rh5_dataflow_joint_ctrl}
\end{figure}

The RH5 humanoid robot uses a hybrid control approach that combines local control loops for low-level motor control and central controllers for high level control as depicted in \rfig{fig:rh5_dataflow_joint_ctrl}.

\subsubsection{Decentralized Actuator-Level Controllers}
In particular, each of the individual actuators is controlled by dedicated electronics placed near the actuator.
On the hardware side, this modular approach facilitates the cabling effort, as it is sufficient to have shared power lines for digital communication to the central controllers. The individual electronics are composed of one or two motor driver boards, a processing board based on a \emph{Xilinx Spartan 6} Field Programmable Grid Array (FPGA), and a board connecting sensors and communication lines. 
In addition, the hardware structure at the control level allows decentralized low-level control, which enables local control loops with low latency.
These local controllers are implemented as a cascade of feedback controllers for motor current, velocity and position, which runs at frequencies of 32 KHz, 4 KHz and 1 KHz, respectively.
Additionally, the local controllers provide feed-forward connections to the high level controllers. This allows to feed-forward velocity and motor current, therefore the amount of feedback can be locally limited to achieve a desired compliant behavior. 
Note that joint position and velocity can be mapped between the independent joint space and actuator space locally, which is also needed for the initialization of the motor's incremental encoder position offset if the absolute position sensor measures the independent joint position.

\subsubsection{Central Electronics for Mid- \& High-Level Control}
A hybrid FPGA / ARM-based system translates and routes status and command messages between the actuators, sensors and a central control PC connected via an Ethernet connection. In order to maximize the transmitted packets to the central control PC while guaranteeing a an upper limit of transmission delay, we implement a routine to synchronize the translation layer to the command messages.
On the control PC, the robot middleware ROCK is used. Software components within this framework act as drivers, which handles the actuator setup and data exchange. It also provides a robot-agnostic interface to the software components implemented in the high level control. The driver components run periodically at a frequency of 1 kHz, resulting in a round trip time of 1 ms.


\section{Analysis and control using DDP}
This section describes the trajectory optimization approach and outlines the simulation and control architecture.    
\label{sec_analysis_and_control_using_DDP}

\subsection{Contact Stability Soft-Constrained DDP}
\subsubsection{Formulation of the Trajectory Optimization Problem} 
Consider a system with discrete-time dynamics as
\begin{equation}\label{eqn:discreteDynamics}
\bx_{i+1}=\bfun(\bx_i,\bu_i), 
\end{equation}
which can be modeled as a generic function $\bfun$ that describes the evolution of the state $\bx\in \myM{R}^n$ from time $i$ to $i+1$, given the control $\bu\in \myM{R}^m$.
The \textit{total cost} $J$ of a trajectory can be written as the sum of running costs $\ell$ and a final cost $\ell_f$ starting from the initial state $\bx_0$ and applying the control sequence $\bu$ along the finite time-horizon:     
\begin{equation}\label{eqn:totalCost}
J(\bx_0, \bu)=\ell_f(\bx_N)+\sum_{i=0}^{N-1}\ell(\bx_i,\bu_i).
\end{equation}
The cost $\ell$ at one discrete time-point (i.e., node) of the optimization depends on the assigned weight $\alpha_c$ and the according cost term $\Phi_c$ as  
\begin{equation}
\ell=\sum_{c=1}^{C}\alpha_c\Phi_c(\bx, \bu). 
\end{equation}
Hence, we write the generic optimal control problem  as
\begin{align}\label{eqn:OCFormulation}
\myM{X}^*,\myM{U}^*= 
&\arg\min_{\mathbf{X},\mathbf{U}} \ell_N(x_N)+\sum_{k=0}^{N-1} \int_{t_k}^{t_k+\Delta t} \ell_k(\bx, \bu)dt,\\ 
&\text{s.t.}\quad\quad\quad \underline{\bu} \leq \bu \leq \bar{\bu},\\ \nonumber
&\quad\quad\quad\quad \underline{\dot{\myM{x}}} = \myM{g}(\bx,\bu) \nonumber
\end{align}
where a complete trajectory $\myM{X},\myM{U}$ is a sequence of states $\myM{X}=\{\bx_0, \bx_1, ..., \bx_N\}$ and control inputs $\myM{U}=\{\bu_0, \bu_1, ..., \bu_N-1\}$ satisfying \cref{eqn:discreteDynamics} and the system dynamics, and $\underline{\bu}$ and $\bar{\bu}$ are the lower and upper torque limits of the system, respectively.

To solve the trajectory optimization problem of \cref{eqn:OCFormulation}, we use the Box-FDDP algorithm \cite{mastalli2020direct}, which is publicly available in the open-source library \textsc{Crocoddyl} \cite{mastalli20crocoddyl}. The Box-FDDP algorithm can compute highly-dynamics motions thanks to its direct-indirect hybridization approach.

\subsubsection{System Dynamics} 
The dynamics of floating base systems is given as:
\begin{equation} \label{eqn:EoMLeggedRobotSurfaceContact}
\myM{M}(\bq)\dot{\myM{v}}+\myM{h}(\bq, \myM{v})=\myM{S}\btau+\sum_{i=1}^{k}\myM{J}_{c_i}^T\myM{\lambda},
\end{equation}
where $\myM{M}$ is the generalized inertia matrix, $\bq \in SE(3) \times \mathbb{R}^n$ are \emph{generalized coordinates}, $\myM{v}$ is the tangent vector, $\myM{S}$ is the actuator selection matrix, $\myM{J}_{c_i}$ is the Jacobian at the location of a contact frame $c_i$ and $\myM{w}_i$ is the \textit{contact wrench} acting on the contact link $i$.  

\subsubsection{Rigid Contact Constraints}
Contacts can be expressed as kinematic constraint on the equation of motion \cref{eqn:EoMLeggedRobotSurfaceContact} as
\begin{equation}
\myM{J}_{c} \dot{\bv} + \dot{\myM{J}}_c \bv = \myM{0}.
\end{equation}
In order to express the holonomic contact constraint $\phi(\bq)=0$ in the acceleration space, it can be differentiated twice. Consequently, the contact condition can be seen as a first order differential-algebraic equation with $\myM{J}_{c}= \begin{bmatrix} \myM{J}_{c_1} & \cdots & \myM{J}_{c_k}\end{bmatrix}$ as a stack of $f$ contact Jacobians.
Finally, the multi-contact dynamics can be expressed as
\begin{equation}\label{eqn:KKTConditions}
\left[\begin{matrix}\myM{M} & \myM{J}^{\top}_c \\{\myM{J}_{c}} & \myM{0}\end{matrix}\right] \left[\begin{matrix} \dot{\bv} \\ -\boldsymbol{\lambda} \end{matrix}\right] = \left[\begin{matrix} \myM{S\tau}-\myM{h} \\ -\dot{\myM{J}}_c \bv\end{matrix}\right].
\end{equation}
For more details about the hybrid optimal control (OC) using this contact dynamics see \cite{budhiraja2018differential}. 

\subsubsection{Optimization Constraints}
We consider constraints of the trajectory optimization problem via a cost-penalization in \cref{eqn:OCFormulation}. Cost terms can either incorporate equality or inequality constraints, which are described in the following.  

In case of equality constraints, an arbitrary task can be formulated as a quadratic regulator term as 
\begin{equation*} 
\Phi_{\text{c}}=\mid\mid \myM{f}(t)-\myM{f}^{\text{ref}}(t)\mid\mid^2_2,
\end{equation*}
where $\myM{f}(t)$ and $\myM{f}^{\text{ref}}$ are actual and reference features, respectively. 
The DDP algorithm utilizes the derivatives of these regulator functions, namely computing the Jacobians and Hessians of the cost functions. 
We use equality constraints for the the CoM tracking ($\Phi_{\text{CoM}}$) and the tracking of the left- and right-foot pose ($\Phi_{\text{foot}}$), respectively.

Equally important for physically consistent trajectory optimization is the consideration of boundaries, such as robot limits and stability constraints. These inequality constraints can be included as penalization term as well. To do so, we use a bounded quadratic term as
\begin{equation}
\Phi_{\text{c}}=
\begin{cases}
\quad\dfrac{1}{2}\myM{r}^T\myM{r} &\mid \underline{\mathbf{r}} > \myM{r} > \bar{\mathbf{r}} \\[10pt]
\quad 0 &\mid \underline{\mathbf{r}} \leq \myM{r} \leq \bar{\mathbf{r}}, 
\end{cases}
\end{equation}
where $\myM{r}$ is the computed residual vector and $\underline{\mathbf{r}}$ and $\bar{\mathbf{r}}$ are the lower and upper bounds, respectively. 
In the scope of our work, we define inequality constraints for joint position and velocity limits ($\Phi_{\text{joints}}$), friction cone constraints ($\Phi_{\text{friction}}$) and center of pressure ($\Phi_{\text{CoP}}$). 

Additional to the described constraints for tasks and physical consistency, we optimize for minimization of the torques ($\Phi_{\text{torques}}$) and regularize the robot posture ($\Phi_{\text{posture}}$).

\subsubsection{Contact Stability}
A key objective in trajectory optimization for legged systems is to ensure a balanced motion that prevents the robot from sliding and falling down. 
We ensure the robot stability by applying the concept of contact wrench cone \cite{caron2015stability}, instead of the widely accepted zero-moment point criterion \cite{vukobratovic1972stability}. 
Note that the latter method is limited due to the assumptions of sufficiently high friction and the existence of one planar contact surface; instead, the former also is suitable for multi-contact OC.   

To this end, we model 6D surface contacts in the OC formulation of \cref{eqn:OCFormulation} with dedicated inequality constraints for unilaterality of the contact forces, Coulomb friction on the resultant force, and center of pressure (CoP) inside the support area: 
\begin{align}\label{eqn:contractWrenchConeReduced}
\begin{split}
\lambda^z &> 0,\\
\mid \lambda^x\mid &\leq \mu \lambda^z,\\
\mid \lambda^y\mid &\leq \mu \lambda^z,\\
\mid X\mid & \geq c_x,\\
\mid Y\mid & \geq c_y.
\end{split}
\end{align}
In \cref{eqn:contractWrenchConeReduced} $\mu$ denotes the static coefficient of friction and models a spatial friction cone, and $c_x$ and $c_y$ denote the position of the CoP with respect to the dimensions $X$ and $Y$ of the rectangular robot feet. This motion planning approach is what we call the \textit{contact stability soft-constrained DDP} \cite{esser2020highly}.

\subsection{Simulation and Control Architecture}
\begin{figure}	
\includegraphics[width=.5\textwidth]{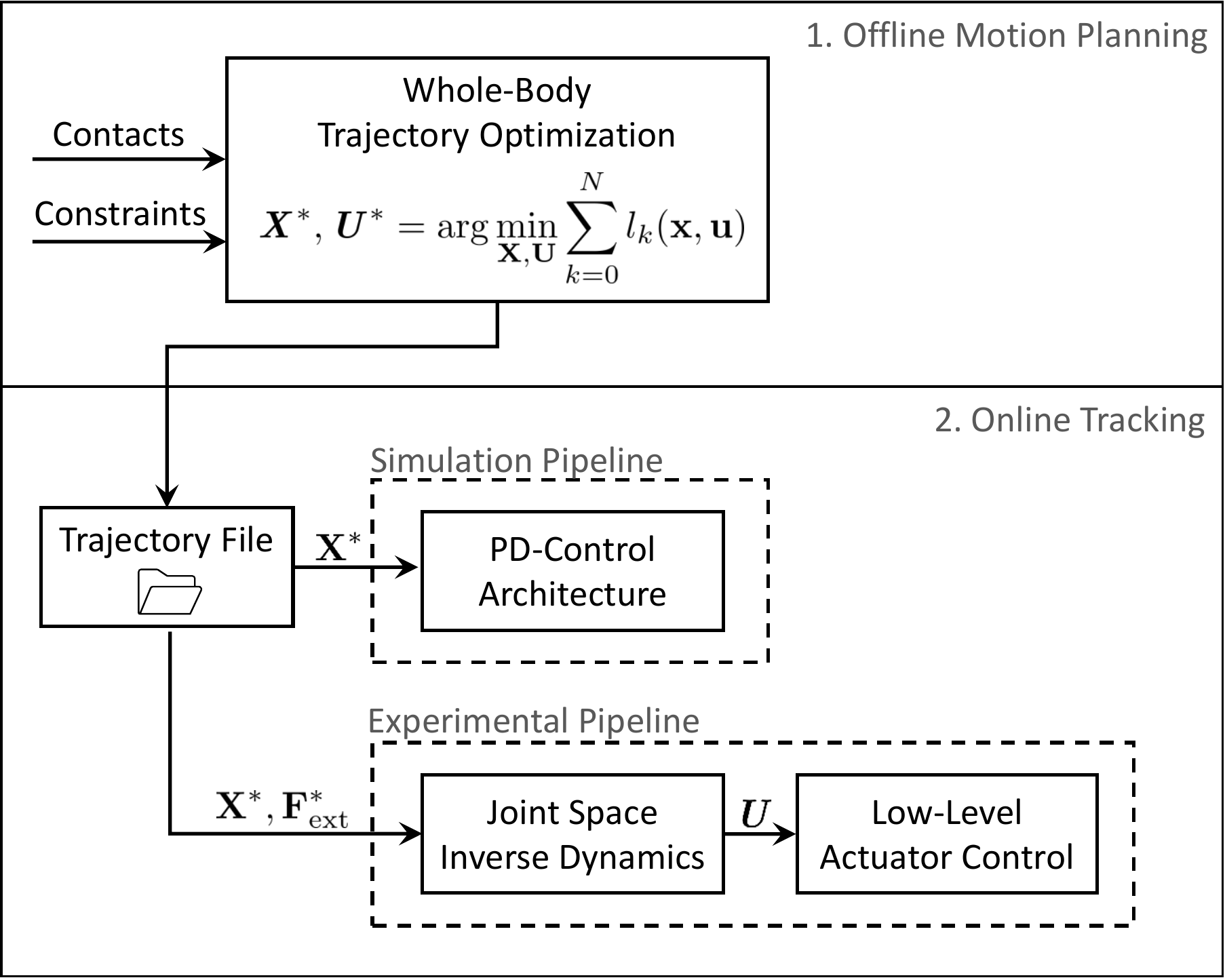}
\caption{
Simulation and experimental pipeline.
}
\label{img:experimentalSetup}
\end{figure}

We track the motions planned with the proposed trajectory optimization approach in real-time with a PD-controller in the PyBullet simulator and with a joint space online stabilization on the real system as depicted in \cref{img:experimentalSetup}. In the following, details on the involved components are provided. 

The contact stability soft-constrained DDP approach computes inherently balanced motions that are concisely captured in an appropriate file. This trajectory file contains the optimal state trajectories $\myM{X}^*$, OC inputs $\myM{U}^*$ and the resulting contact wrenches $\myM{F}_{\text{ ext}}^*$ acting on the feet. The trajectories are interpolated to 1 kHz using cubic splines in order to ensure smoothness. 
The planned motions are computed based on a tree type robot model. For dynamic real-time control, this simplified model turns out to be sufficient, although the accuracy is reduced \cite{kumar2019model}. Nevertheless, the problem remains on transforming the results from the independent joint space, to the actuation space. We use the modular software framework HyRoDyn (Hybrid Robot Dynamics)~\cite{kumar2020analytical} to map the trajectories generated for the serialized robot model to compute the forces of the respective linear actuators.
Low-level actuator controllers compensate deviations from the reference trajectories. Analogously to the simulation pipeline, this real-time control approach uses a cascaded feedback of position, velocity and an additional current control loop. 
\section{Results and discussions}
\label{sec_results_discussions}
This section presents the evaluation of the robot design, simulation results and first experimental trials. 

\subsection{Evaluation of Robot Design}
We evaluated the RH5 humanoid design by performing a wide range of complex motions. The motivation is to form a basis of decision-making for future design iterations that allow us to perform such tasks. \cref{tab:motionCharacteristics} provides details on the performed motions and \cref{tab:systemLimits} summarizes the results. 

\subsubsection{Dynamic Walking Variants}
\begin{figure}
\centering	
\includegraphics[width=.5\textwidth]{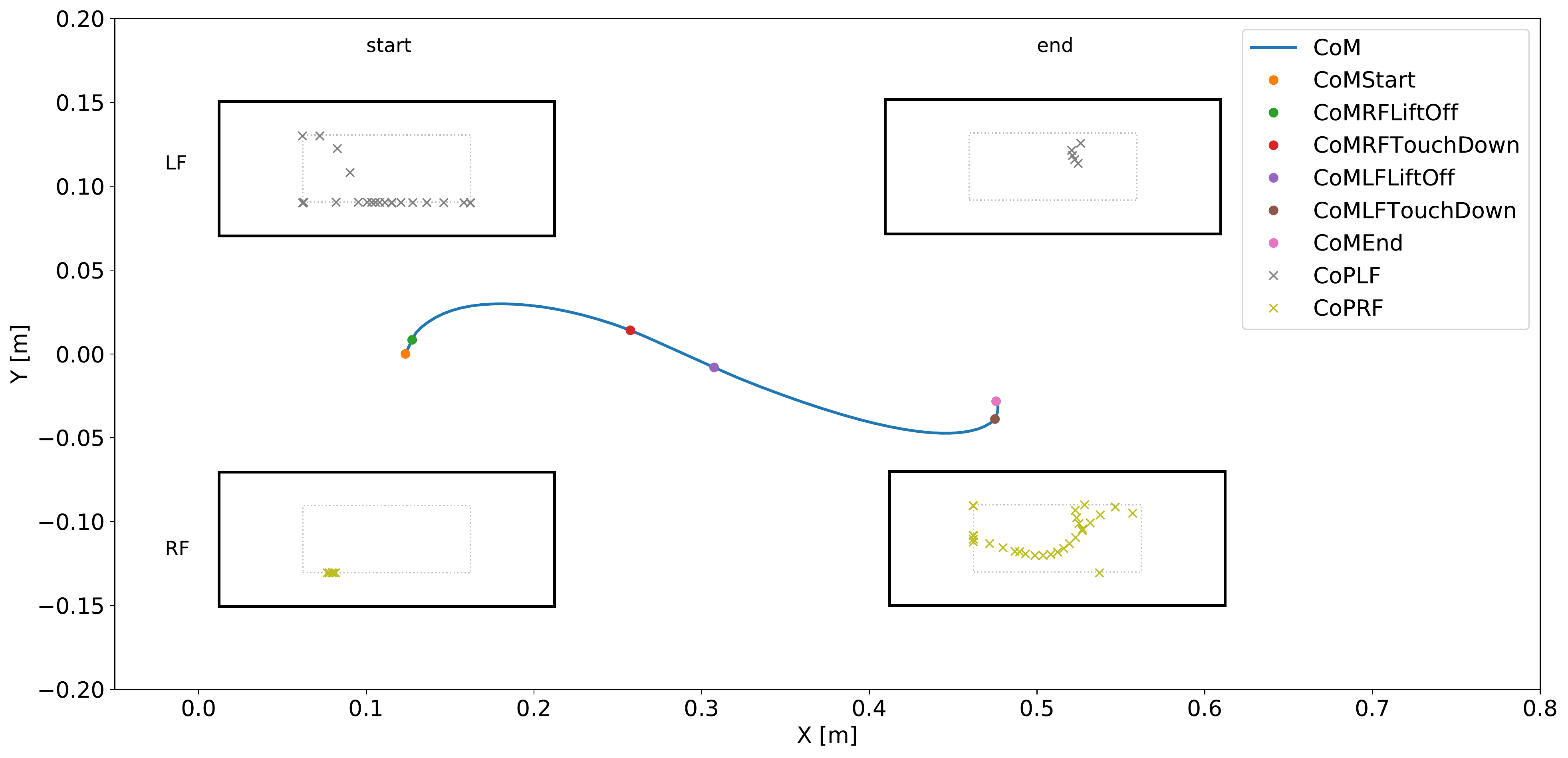}
\caption{Dynamic walking gait computed with the contact stability soft-constrained DDP. The motion is inherently balanced, since the CoPs (crosses) for both feet (LF, RF) remain inside the desired CoP region of 50\% foot coverage (dashed lines).}
\label{fig:walkCoP}
\end{figure}

We study efficient motions for dynamic walking gaits with high velocities. To this end, we apply the proposed approach of contact stability soft-constrained DDP, where the CoP of each foot is constrained. By this, the solver is enabled to find an optimal, dynamic CoM shifting along with the requested contact stability constraints. 
\cref{fig:walkCoP} shows this approach yields dynamically balanced walking motions where the CoP of each foot (crosses) in contact stays within a predefined range. Following our motion planning approach, we observed that often, for speeds greater than 0.35 m/s, the solver needs to be initialized with a predefined CoM trajectory in order to find a feasible solution, as done in \cite{budhiraja2018differential}.

\begin{figure}
\centering	
\includegraphics[width=.5\textwidth]{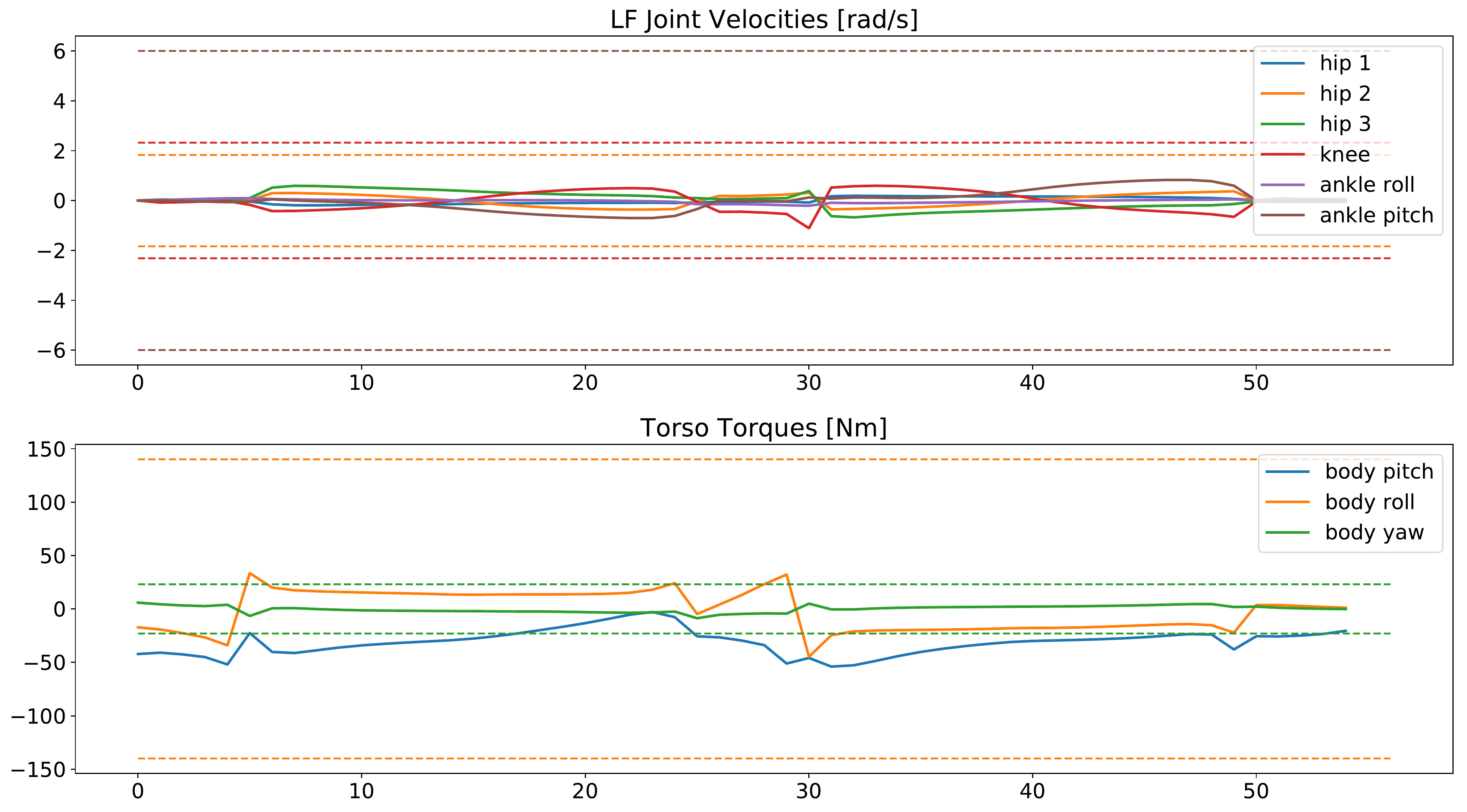}
\caption{Optimal solution of the dynamic walking gait with 5 kg weights in each hand within the robot's velocity and torque limits (dashed lines).}
\label{fig:walkWeights_jointSpace}
\end{figure}

\paragraph{Walking with weights (5kg Bars)}
We evaluated the capabilities of the RH5 robot to perform a dynamic walking gait at 0.35 m/s while carrying 5kg aluminum weights in each hand (see \cref{fig:walkWeights_snaps}). A natural CoM shifting emerges resulting from the inequality constraints for the CoP of each foot. \cref{fig:walkWeights_jointSpace} shows that the found optimal solution is within the joint position and velocity limits as well as torque limits. 


\paragraph{Walking with high speed (1 m/s)}
In order to analyze the limits of the RH5 humanoid, we successfully performed a fast dynamic walking gait at 1 m/s with a predefined CoM trajectory (see \cref{tab:motionCharacteristics}). Also for this dynamic walking gait, the OC solver found a feasible solution within the robot limits, proofing for the versatility of the RH5 robot design.

\subsubsection{Squatting with Weights}
\begin{figure}
\begin{subfigure}{.16\textwidth}
	\includegraphics[width=.95\linewidth]{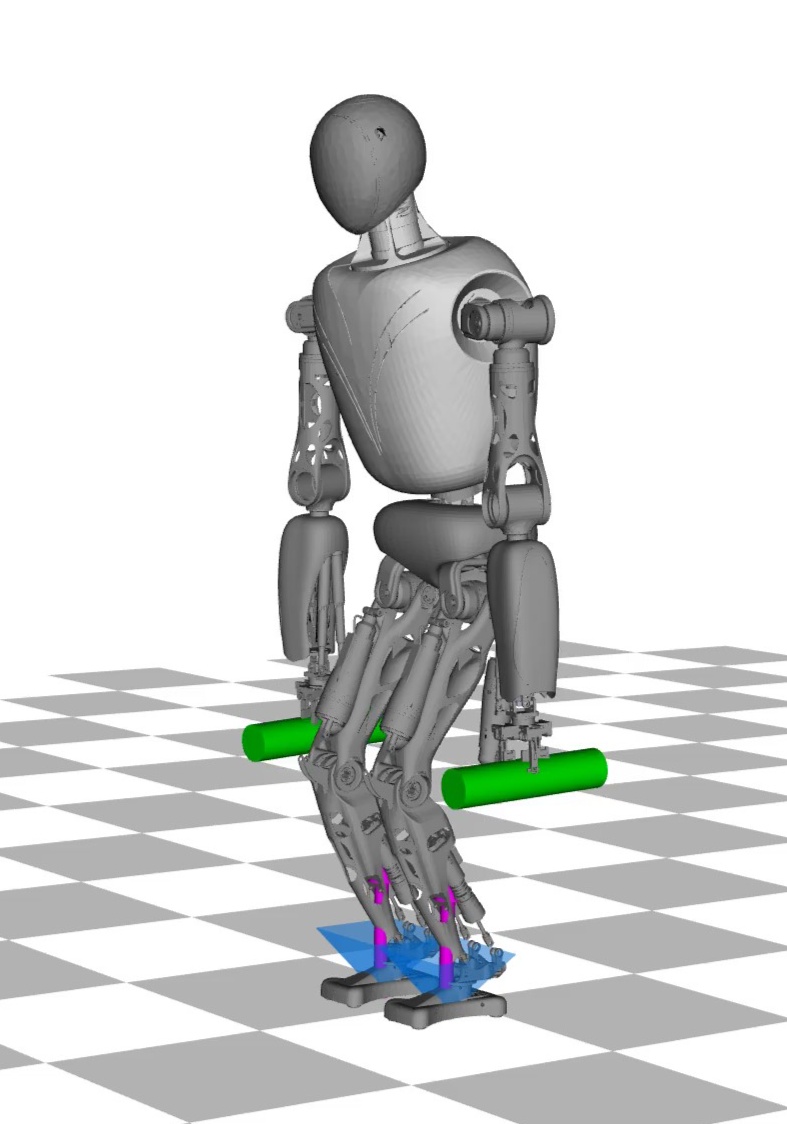}
	\caption{}
\end{subfigure}%
\begin{subfigure}{.16\textwidth}
	\includegraphics[width=.95\linewidth]{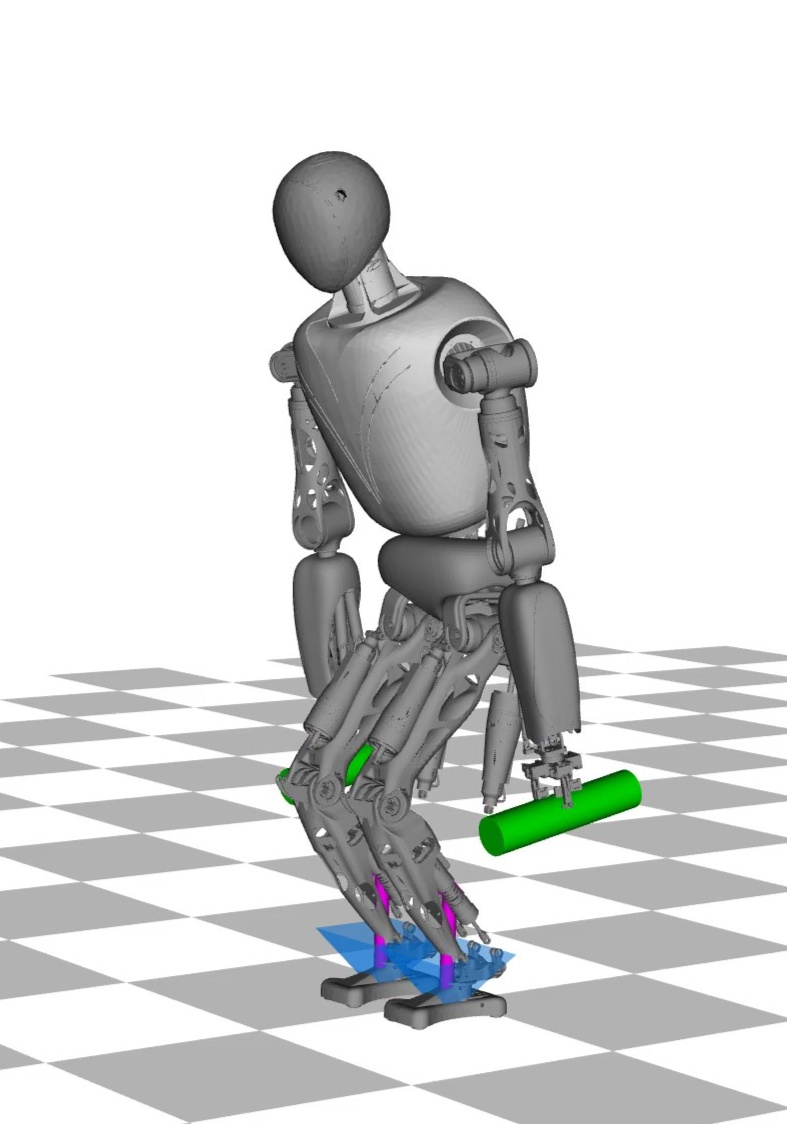}
	\caption{}
\end{subfigure}%
\begin{subfigure}{.16\textwidth}
	\includegraphics[width=.95\linewidth]{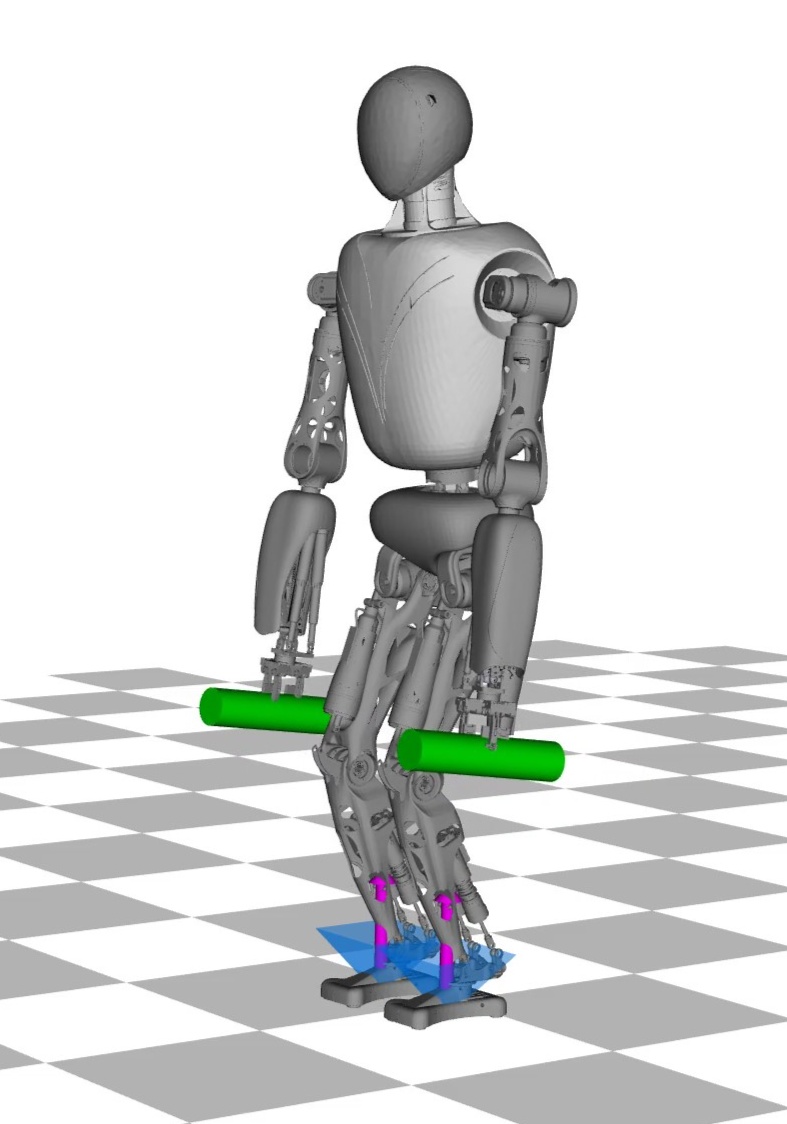}
	\caption{}
\end{subfigure}%
\caption{Squatting with 5kg aluminum bars in each hand.}
\label{fig:squatting_snaps}
\end{figure}

As for fast dynamic walking, we analyzed a sequence of dynamic squatting movements with a predefined CoM range of 20 cm (see \cref{fig:squatting_snaps}). We also found that the joint position, velocity and torque limits were satisfied.  

\subsubsection{Jumping Variants}
We analyzed the limits of the system design by performing highly-dynamic jumps. 

\paragraph{Vertical jumping}
Although the RH5 robot has been designed for walking motions and not highly-dynamic ones, vertical jumps with a height of 1 cm can be performed within valid ranges for joint position, velocity and torques. For the case of a 10 cm jump the joint position and torques limits are within the limits. However, velocity peaks at the take-off exceed the limits of the body pitch and knee joints by a factor of two and four, respectively. This effect is plausible, since both the knee as well as the torso swing are essential for a jump. We deployed a heuristic approach to identify the minimal design improvement by scaling the critical joint limits step by step until a feasible solution is found. For the 10 cm vertical jump we found that an optimal solution is found by scaling only the knee joint velocity limits of the robot by a factor of 3. 

\begin{figure}
\begin{subfigure}{.25\textwidth}
	\includegraphics[width=.9\linewidth]{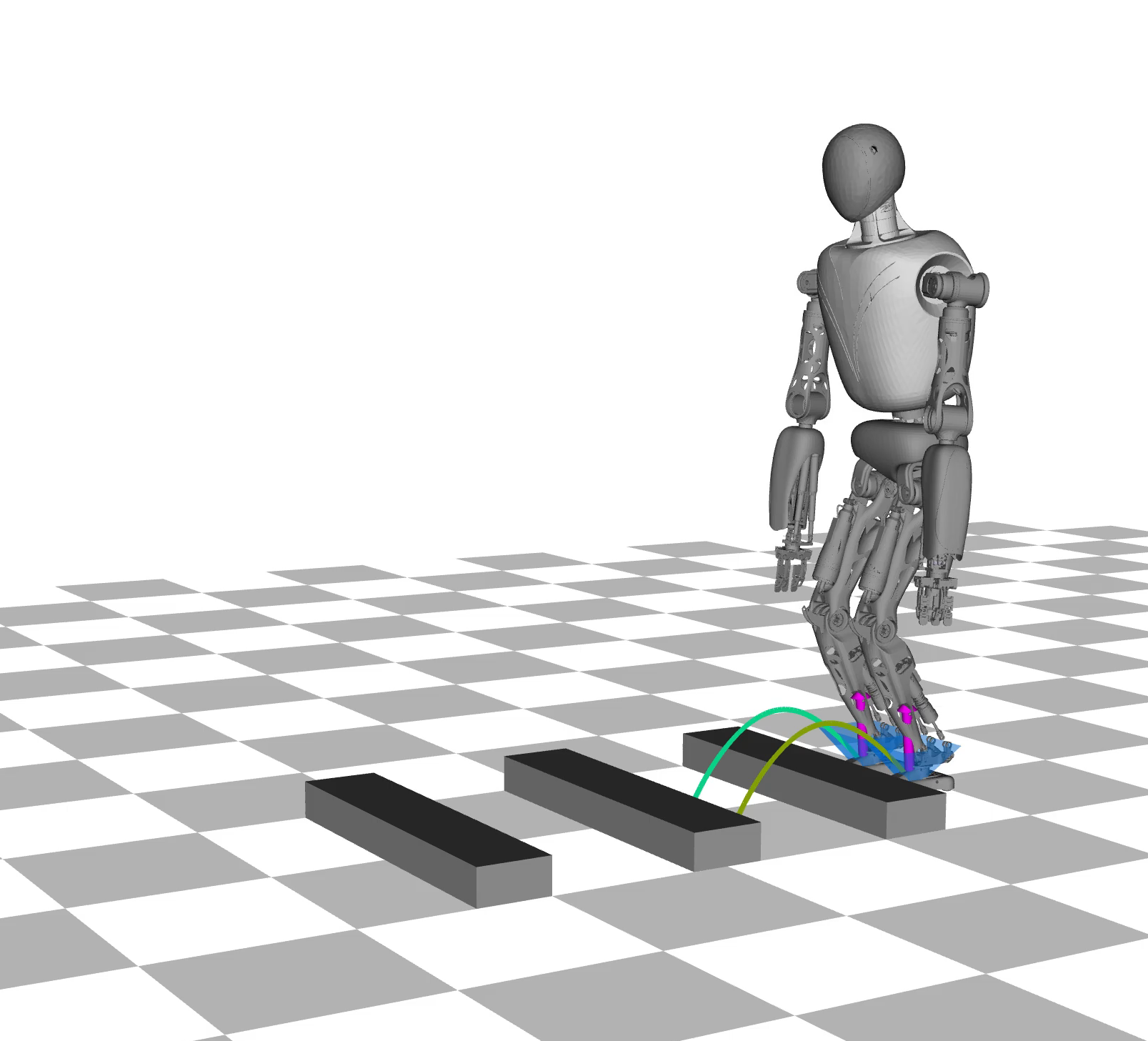}
	\caption{}
	\end{subfigure}%
\begin{subfigure}{.25\textwidth}
	\includegraphics[width=.9\linewidth]{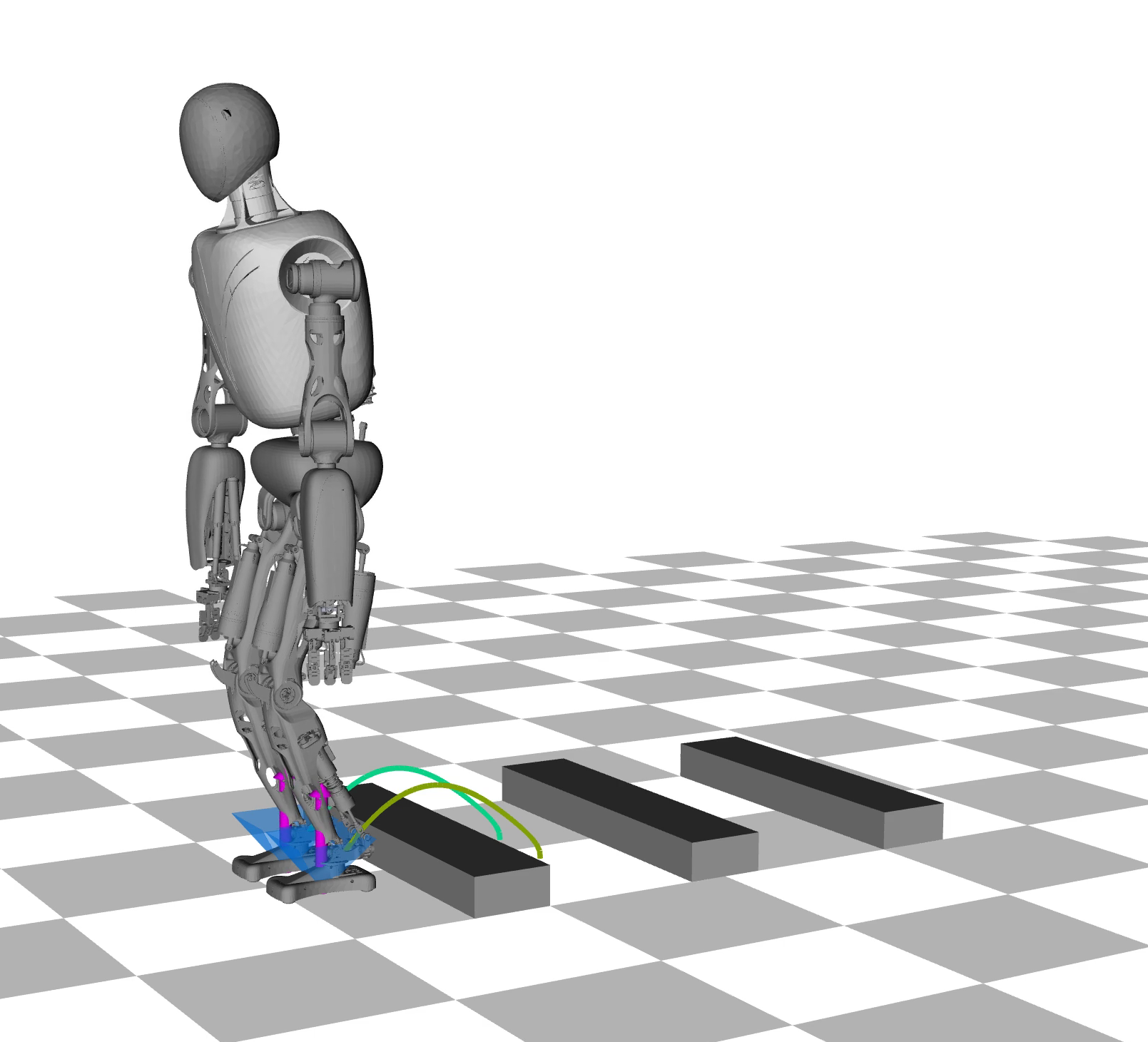}
	\caption{}
\end{subfigure}
\caption{Sequence of challenging jumps over obstacles.}
\label{fig:jumpObstacles_Snaps}
\end{figure}

\paragraph{Jumping over multiple obstacles}
Finally, investigated a more challenging jumping sequence over obstacles (see \cref{fig:jumpObstacles_Snaps}). Since the humanoid was not designed for such tasks, neither joint velocity nor torque limits can be satisfied. Further details on the formulation of the OC problems, used optimization constraints, extracted design guidelines and videos are provided in \cite{esser2020highly}. 

\begin{table*}[t]
\centering
\caption{Characteristics and applied optimization constraints for a wide range of dynamic motions.}
\begin{tabular}{l|cccc|ccccccc}
\hline
& \multicolumn{4}{c|}{\textbf{Motion Characteristics}} & \multicolumn{7}{c}{\textbf{Optimization Constraints}} \\ \hline
& Length & Height & Total time & Step size & \multicolumn{2}{c}{Tasks} & \multicolumn{2}{c}{Stability} & Limits & \multicolumn{2}{c}{Regularization} \\ \hline
& & & & & $\Phi_{\text{foot}}$ & $\Phi_{\text{CoM}}$ & $\Phi_{\text{friction}}$ & $\Phi_{\text{CoP}}$ & $\Phi_{\text{joint}}$ & $\Phi_{\text{posture}}$ & $\Phi_{\text{torque}}$\\ \hline
Dynamic walking with 5kg weights & 0.5 m & 0.05 m & 1.5 s & 0.03 s & \ding{53} & & \ding{53} & \ding{53} & \ding{53} & \ding{53} & \ding{53} \\
Fast dynamic walking (1 m/s) & 0.7 m & 0.1 m & 0.7 s & 0.03 s & \ding{53} & \ding{53} & & & \ding{53} & \ding{53} & \ding{53} \\ 
Squatting with 5kg weights & -- & 0.2 m & 2 s & 0.03 s &  \ding{53} & \ding{53} & \ding{53} & \ding{53} & \ding{53} & \ding{53} & \ding{53}\\
Vertical jump ($h=$ 0.01 m) & --  & 0.01 m & 0.9 s & 0.01 s & \ding{53} & & \ding{53} & \ding{53} & \ding{53} & \ding{53} & \ding{53} \\
Vertical jump ($h=$ 0.1 m) & -- & 0.1 m & 0.9 s & 0.01 s & \ding{53} & & \ding{53} & \ding{53} & \ding{53} & \ding{53} & \ding{53}\\
Jumps over obstacles & 0.6 m & 0.25 m & 2.7 s & 0.01 s & \ding{53} & & \ding{53} & \ding{53} &\ding{53} & & \ding{53} \\ \hline
\end{tabular}
\label{tab:motionCharacteristics}
\end{table*}

\begin{table}[t]
\centering
\caption{Capabilities of the RH5 humanoid to perform a wide range of motions respecting the hardware limits.
}
\begin{tabular}{lccc}
\hline
Experiment & Pos. Lim. & Torque Lim. & Vel. Lim.\\ \hline
Walk with 5kg weights & \greencheckmark  & \greencheckmark & \greencheckmark \\
Dynamic walk (1 m/s) & \greencheckmark  & \greencheckmark & \greencheckmark \\
Squats with 5kg weights & \greencheckmark  & \greencheckmark & \greencheckmark \\
Vertical jump ($h=$ 0.01 m) & \greencheckmark  & \greencheckmark & \greencheckmark \\
Vertical jump ($h=$ 0.1 m) & \greencheckmark & \greencheckmark & \redxmark$_3$  \\
Forward obstacle jumps & \greencheckmark  & \redxmark$_5$ & \redxmark$_7$ \\ \hline
\end{tabular}
\label{tab:systemLimits}
\end{table}

\subsection{Simulation Results}

\begin{figure}
\includegraphics[width=1\linewidth]{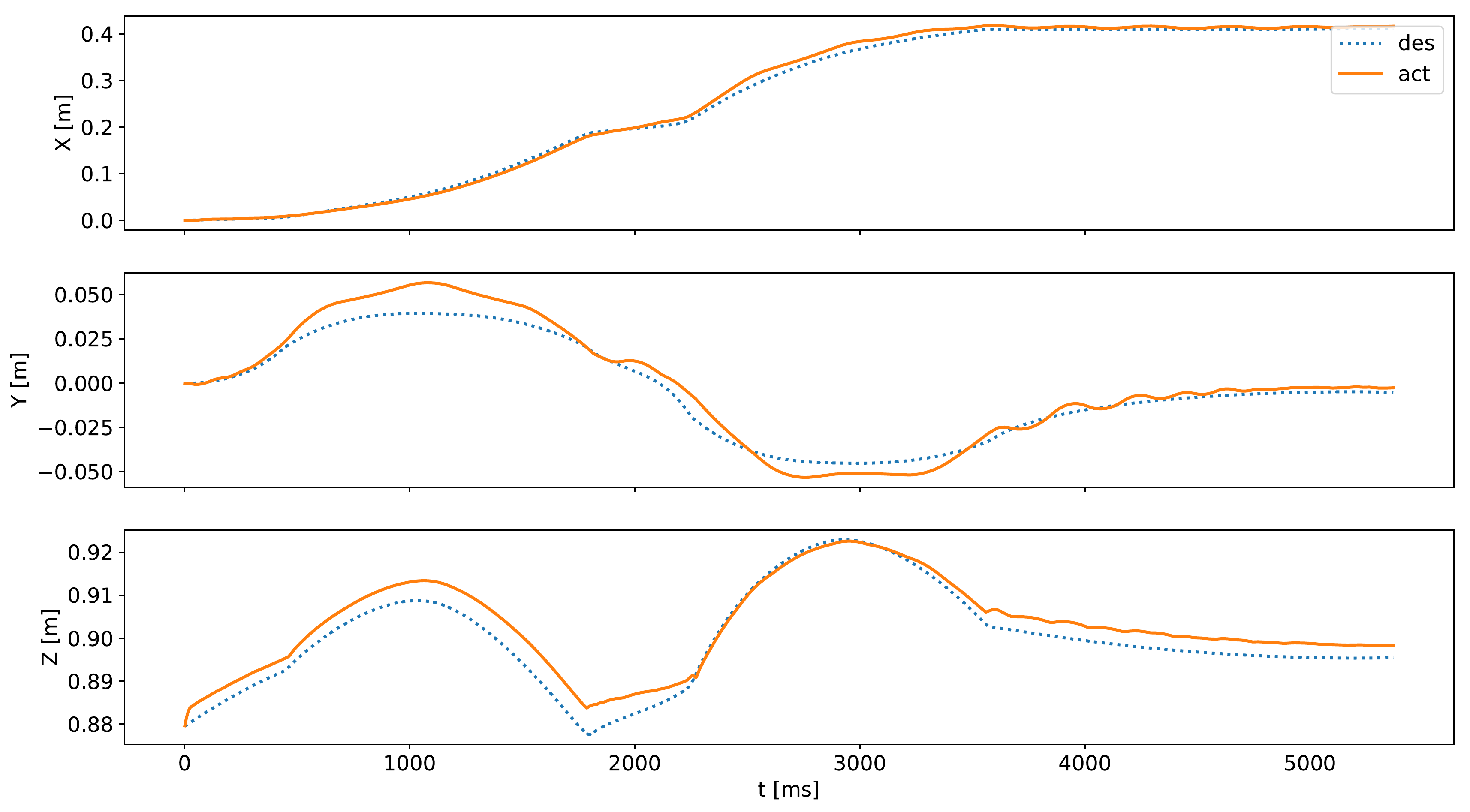}
\caption{Motion of the floating base resulting from joint level control for the dynamic walking gait. 
}
\label{fig:walkDynamic_pybullet}
\end{figure}

We proved the stability of the optimized dynamic walking motion in the PyBullet simulator using a joint space PD controller.
\cref{fig:walkDynamic_pybullet} monitors the optimized motion of the uncontrolled floating base. As can be seen, the floating base deviates about $\pm$ 10 mm in x- and y-direction as well as $+$ 5 mm in z-direction. 
The motions turn out to be inherently balanced due to the proposed contact stability soft-constrained DDP approach. Hence, our trajectories did not require a dedicated online stabilizer, in contrast to the work of \cite{giraud2020motion}, to generate a physically consistent motion. 

\subsection{Experimental Trials}
\begin{figure}
\begin{subfigure}{.1\textwidth}
	\includegraphics[width=.95\linewidth]{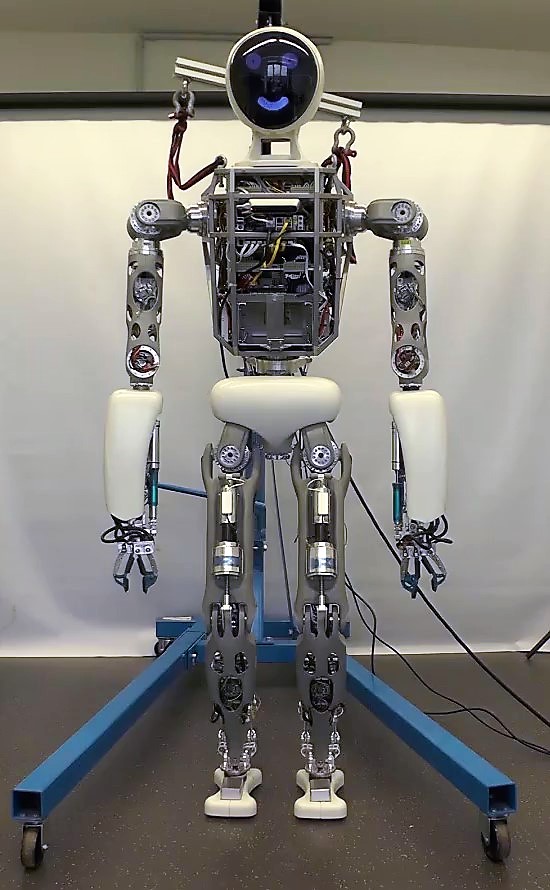}
	\caption{}
	\end{subfigure}%
\begin{subfigure}{.1\textwidth}
	\includegraphics[width=.95\linewidth]{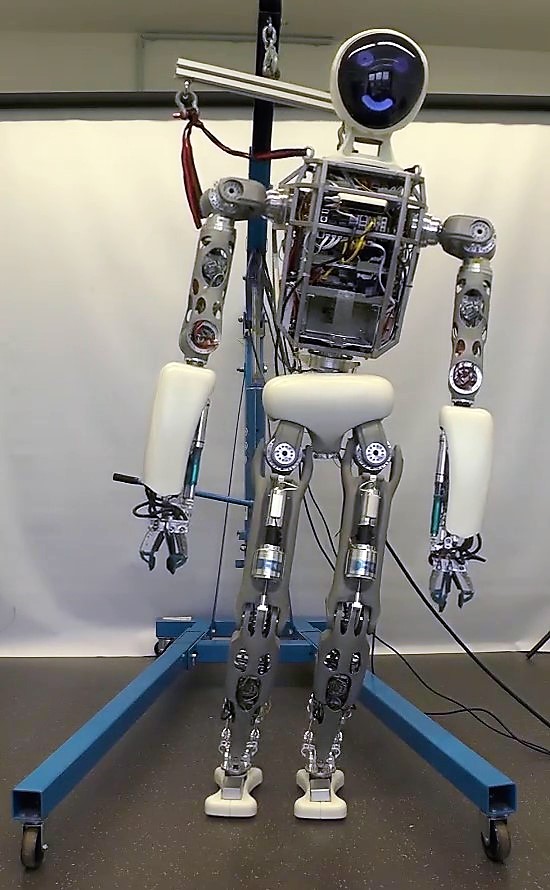}
	\caption{}
\end{subfigure}%
\begin{subfigure}{.1\textwidth}
	\includegraphics[width=.95\linewidth]{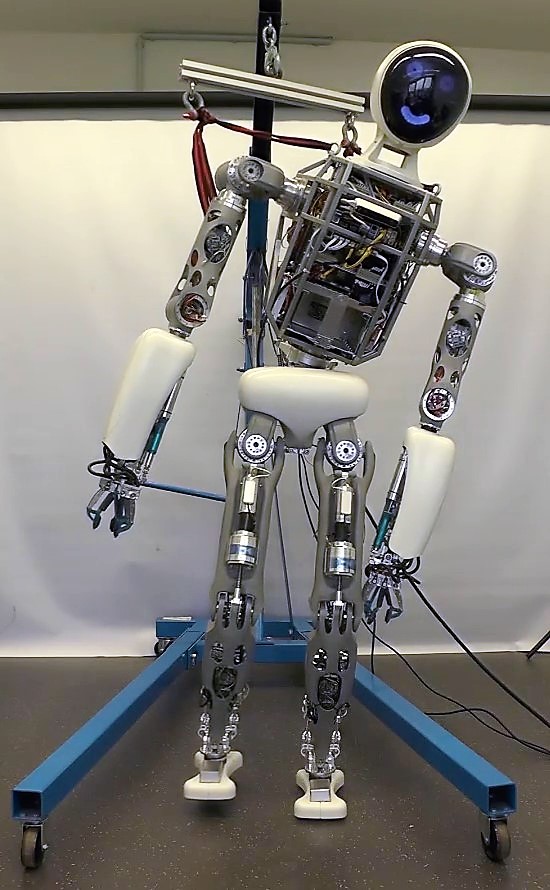}
	\caption{}
	\end{subfigure}%
\begin{subfigure}{.1\textwidth}
	\includegraphics[width=.95\linewidth]{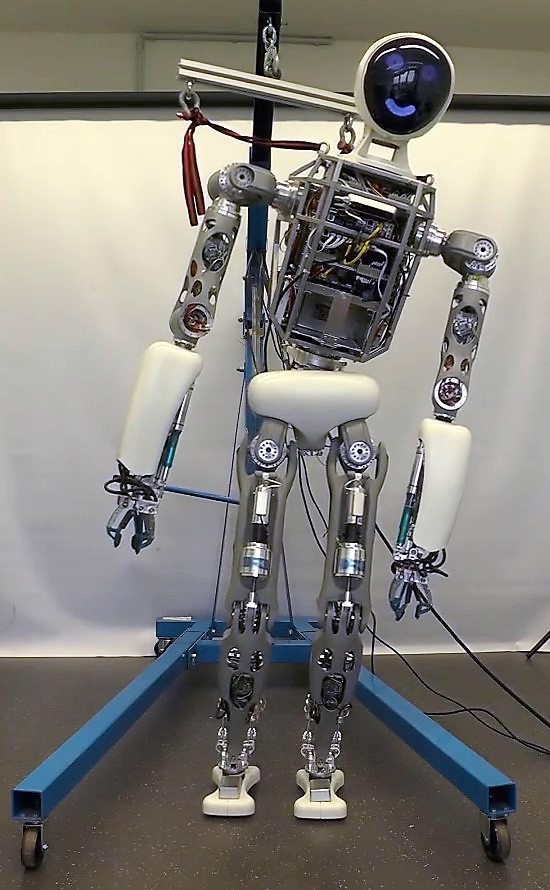}
	\caption{}
\end{subfigure}%
\begin{subfigure}{.1\textwidth}
	\includegraphics[width=.95\linewidth]{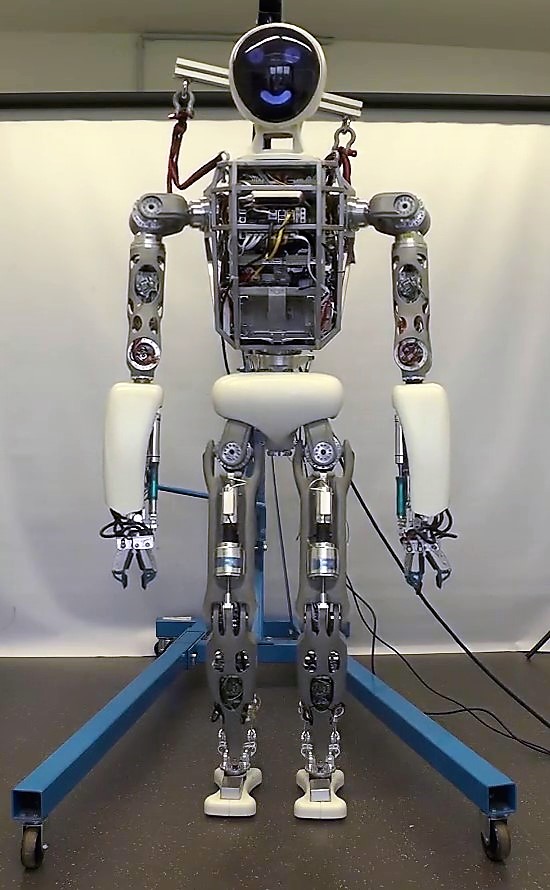}
	\caption{}
	\end{subfigure}%
\caption{Experiment I: one-leg balancing from (a) an initial pose, (b) CoM shift above the LF, (c) lifting the RF up and (d) down and (e) recovering to the initial pose.
} 
\label{exp:balancingSnaps}
\end{figure}

\begin{figure}
\begin{subfigure}{.1\textwidth}
	\includegraphics[width=.95\linewidth]{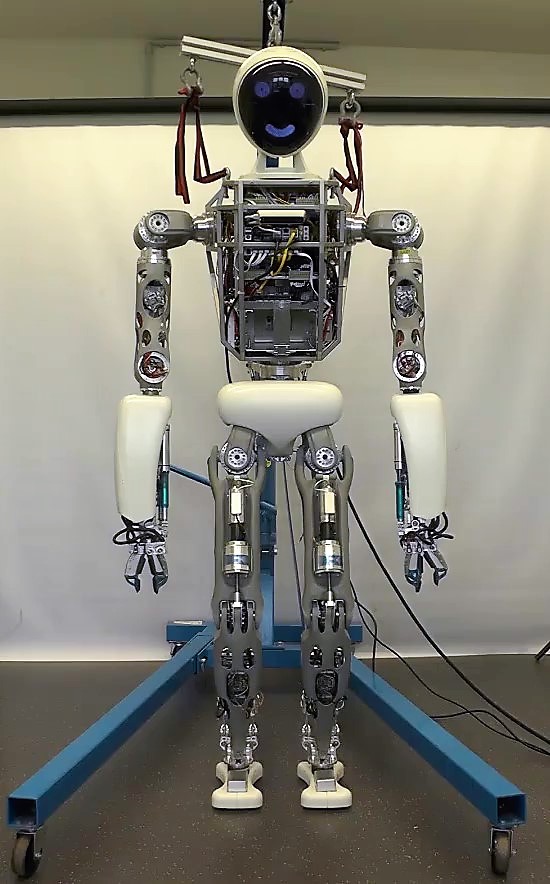}
	\caption{}
	\end{subfigure}%
\begin{subfigure}{.1\textwidth}
	\includegraphics[width=.95\linewidth]{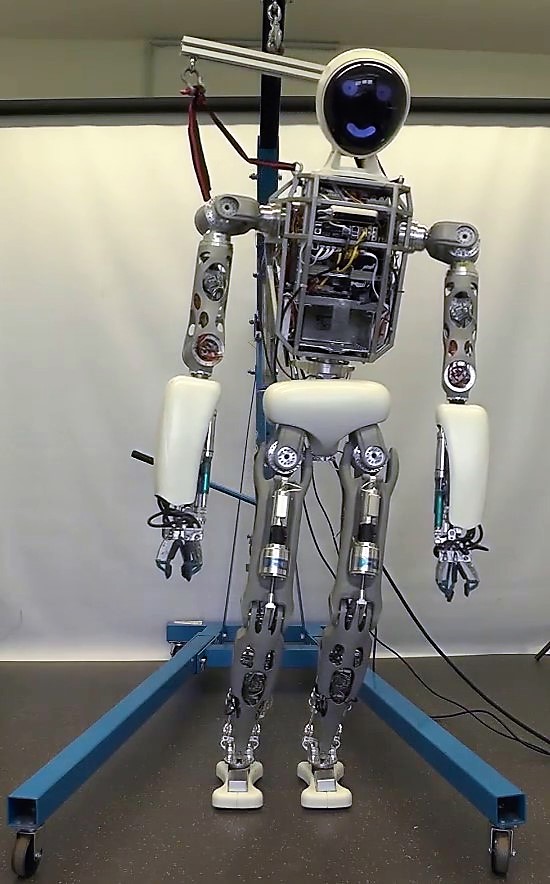}
	\caption{}
\end{subfigure}%
\begin{subfigure}{.1\textwidth}
	\includegraphics[width=.95\linewidth]{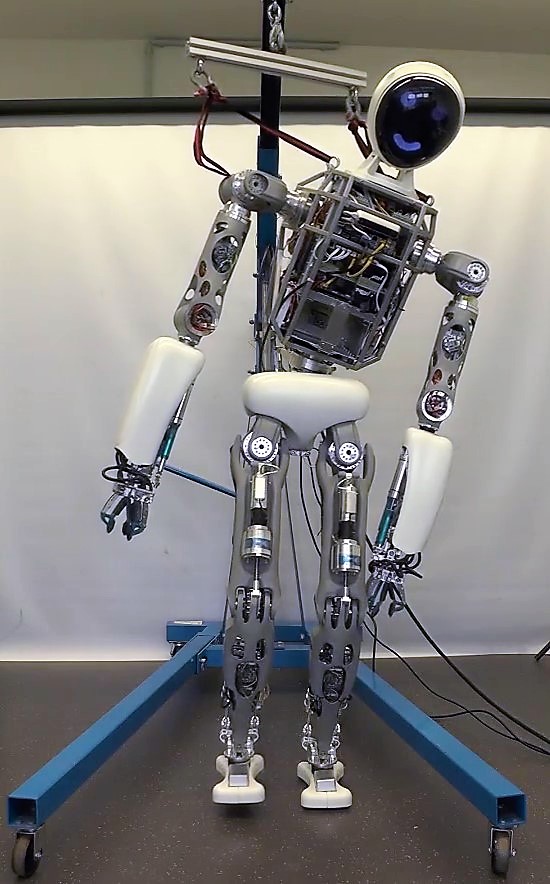}
	\caption{}
	\end{subfigure}%
\begin{subfigure}{.1\textwidth}
	\includegraphics[width=.95\linewidth]{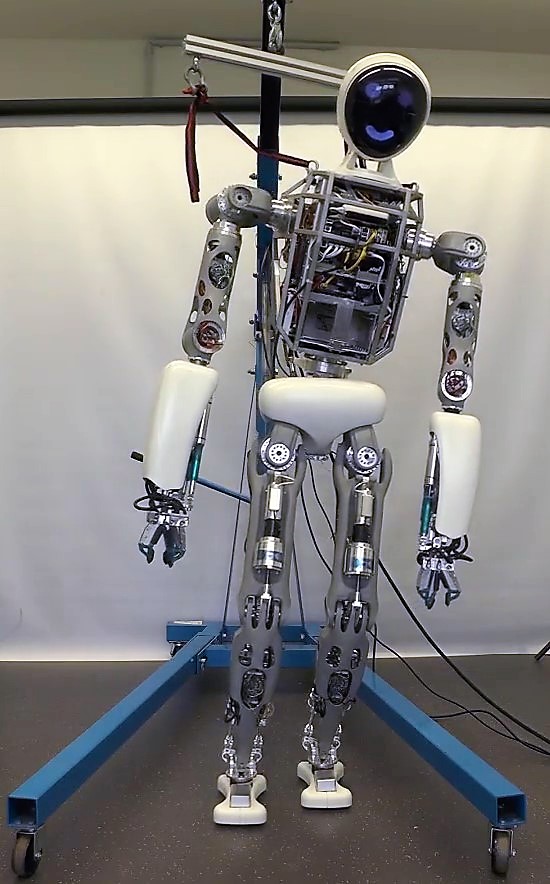}
	\caption{}
\end{subfigure}%
\begin{subfigure}{.1\textwidth}
	\includegraphics[width=.95\linewidth]{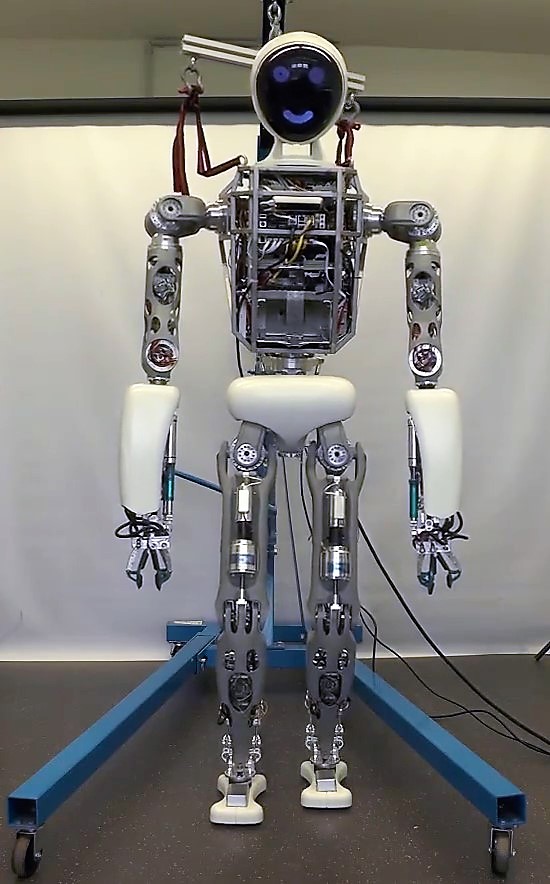}
	\caption{}
	\end{subfigure}%
\caption{Experiment II: static stepping motion from (a) an initial pose, (b) CoM shift above the LF, (c,d) performing a right step and (e) shifting the CoM to the center of the SP.}
\label{exp:staticWalkingSnaps}
\end{figure}

\begin{figure}
\begin{subfigure}{.25\textwidth}
	\includegraphics[page=1, width=.95\linewidth]{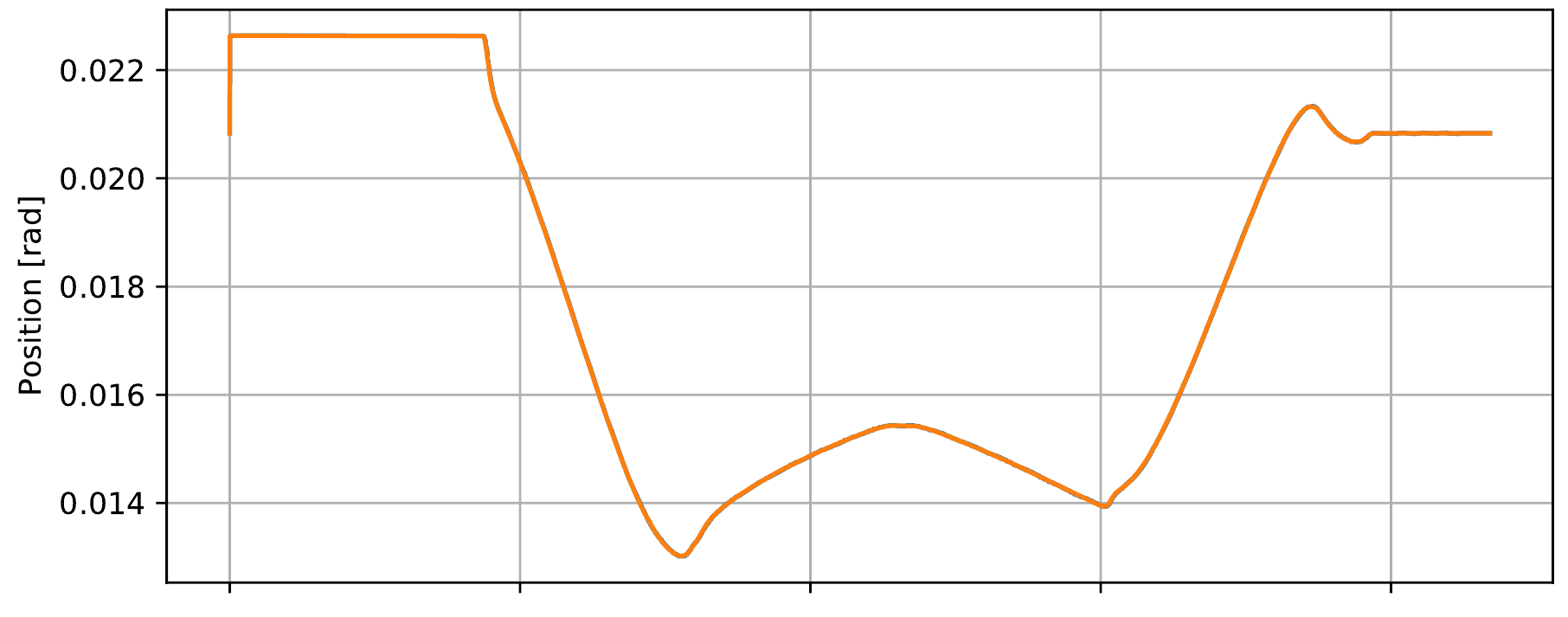}
	\caption{LLAnkleAct1}	
	\end{subfigure}%
\begin{subfigure}{.25\textwidth}
	\includegraphics[page=2, width=.95\linewidth]{experiments/balancing/ankleTrack}
	\caption{LLAnkleAct2}	
\end{subfigure}%

\begin{subfigure}{.25\textwidth}
	\includegraphics[page=3, width=.95\linewidth]{experiments/balancing/ankleTrack}
	\caption{LLAnkleRoll}	
	\end{subfigure}%
\begin{subfigure}{.25\textwidth}
	\includegraphics[page=4, width=.95\linewidth]{experiments/balancing/ankleTrack}
	\caption{LLAnklePitch}	
\end{subfigure}%
\caption{
Tracking performance for the one-leg balancing experiment in actuators (a,b) and independent joints (c,d). 
}
\label{exp:BalancingTracking}
\end{figure}

\begin{figure}
\begin{subfigure}{.1\textwidth}
	\includegraphics[width=.9\linewidth]{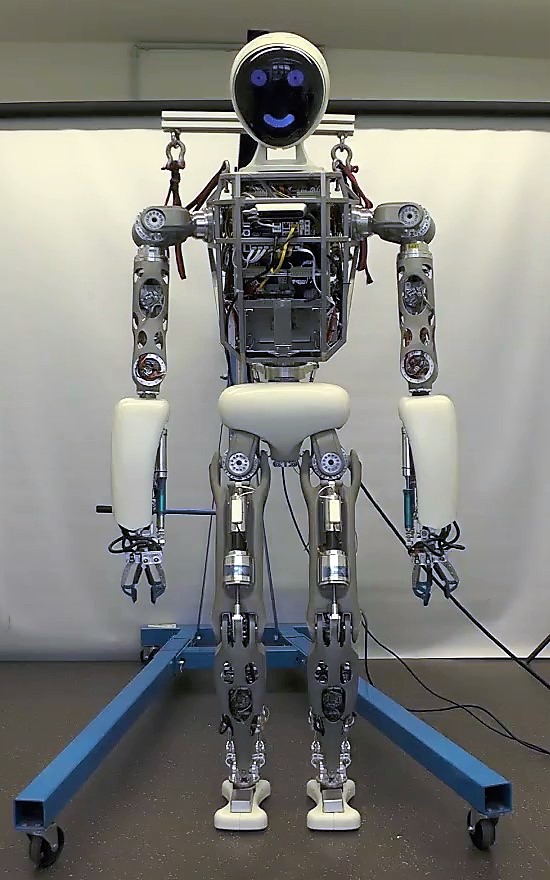}
	\caption{}
	\end{subfigure}%
\begin{subfigure}{.1\textwidth}
	\includegraphics[width=.9\linewidth]{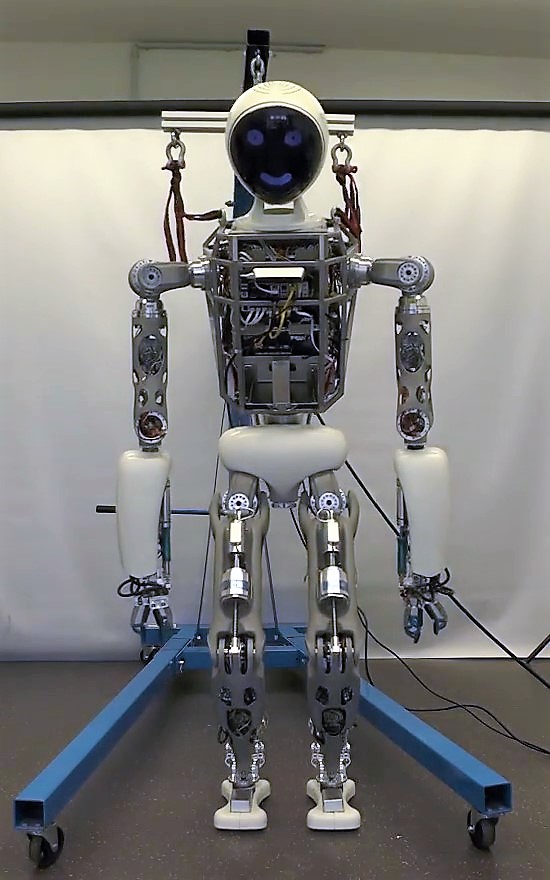}
	\caption{}
\end{subfigure}%
\begin{subfigure}{.1\textwidth}
	\includegraphics[width=.9\linewidth]{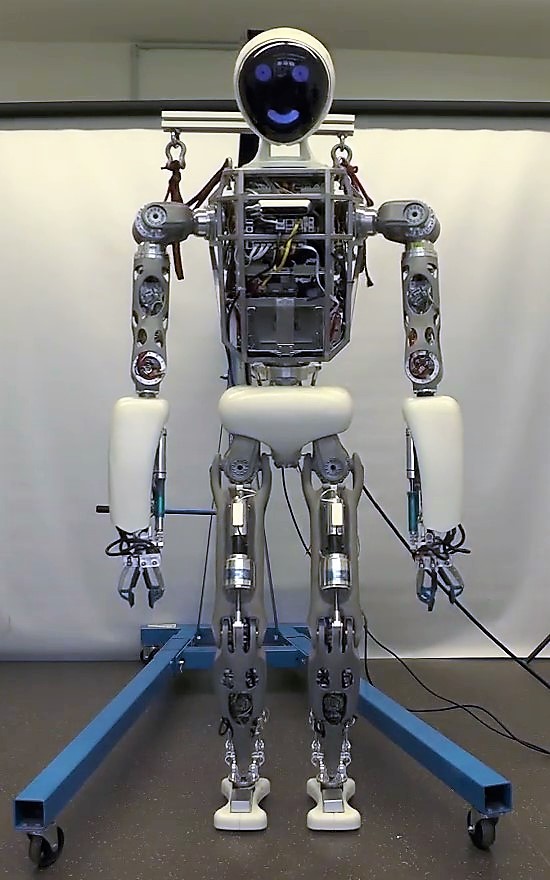}
	\caption{}
	\end{subfigure}%
\begin{subfigure}{.1\textwidth}
	\includegraphics[width=.9\linewidth]{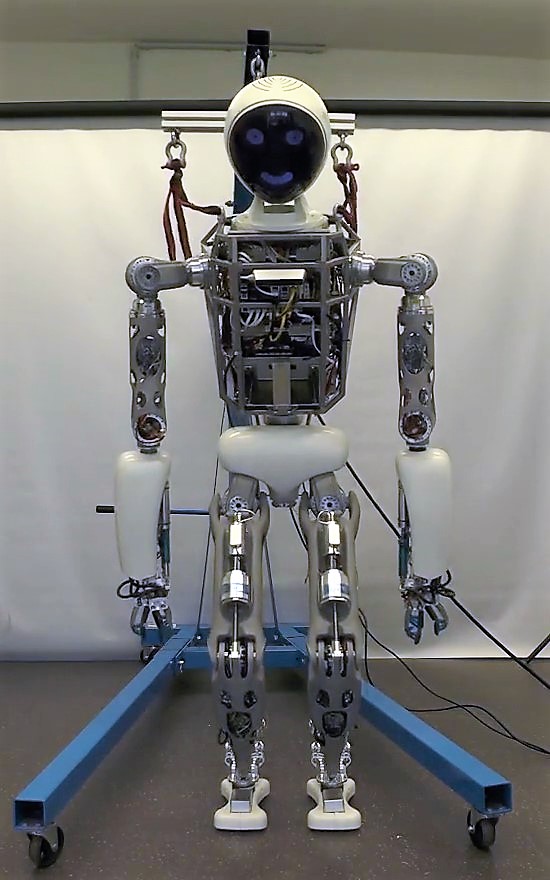}
	\caption{}
\end{subfigure}%
\begin{subfigure}{.1\textwidth}
	\includegraphics[width=.9\linewidth]{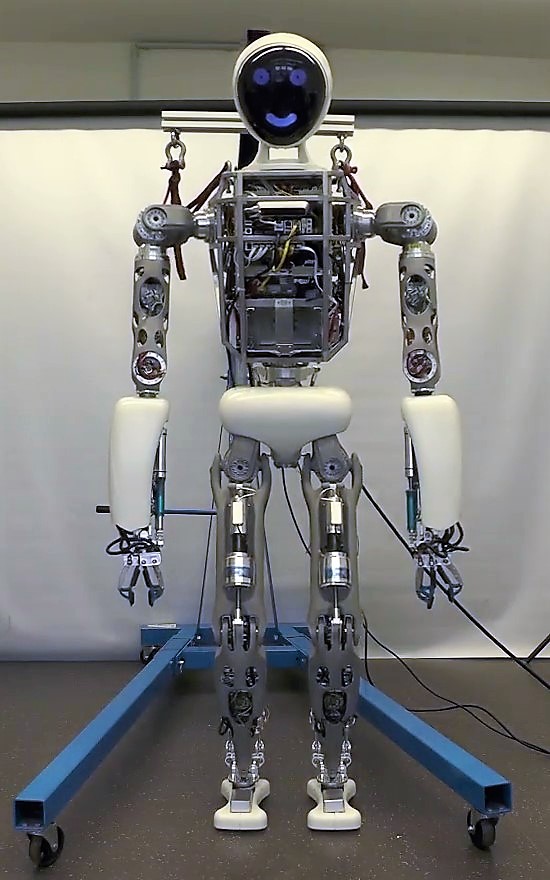}
	\caption{}
	\end{subfigure}%
\caption{Experiment III: sequence of fast squats from (a) an initial pose over, (b,d,f) descending the CoM by 15 cm and (c,e,g) recovering to the initial pose.}
\label{exp:squatSnaps}
\end{figure}

We conducted three experiments with increasing level of difficulty. The goal of the first experiment is to test the ability of the controller to track a slow balancing task. The quasi-static motion consists of five phases as visualized in \cref{exp:balancingSnaps}.
The second experiment deals with a stabilization of a static stepping motion (see \cref{exp:staticWalkingSnaps}). The objective of this test is to analyze the effect of more difficult swing-leg motions, a step sequence of two steps and the effect of impacts.
The objective of the third experiment is to evaluate the tracking performance in the context of a dynamic motion. In contrast to the first two motions, the fast squatting experiment (see \cref{exp:squatSnaps}) involves dynamic forces acting on the robot resulting from a fast vertical base movement in the range of 15 cm within two seconds.
Overall, the three planned motions could be stabilized with good accuracy by the controller on the real system. 
\cref{exp:BalancingTracking} shows the tracking performance for the one-leg balancing experiment. The control architecture allows following the computed reference trajectory closely, both in actuator space (a,b) and independent joint space (c,d). 
This precise tracking is achieved with high-gain joint space control, which allows a quick compensation of position differences that comes at the cost of lost compliance in the joints.
The impact phase turned out to be the main problem for the walking experiments. This is reasonable since the utilized control approach only compensates for errors in joint space, while errors in task space can arise quickly and are not compensated.

%

\section{Conclusion}
\label{sec_conclusion}
This paper presented the design and analysis of a novel series--parallel hybrid humanoid robot named RH5 which has a lightweight design and good dynamic characteristics. 
We see large potential in using DDP-based whole-body TO to evaluate the capabilities of humanoid robots. 
The preliminary experiments indicate that the proposed planning approach efficiently generates physically consistent motions for the RH5 humanoid robot.
Future work includes experiments with online stabilization to realize heavy-duty tasks with the real system. We also plan to address the resolution of internal closed loops along with the holonomic constraints imposed by the contacts within the DDP formulation. 



%


\bibliographystyle{IEEEtran}
\bibliography{references}

\end{document}